\documentclass[10pt,twocolumn,letterpaper]{article}

\usepackage{iccv}
\usepackage{times}
\usepackage{epsfig}
\usepackage{graphicx}
\usepackage{amsmath}
\usepackage{amssymb}

\usepackage{subfigure}
\usepackage{accsupp}
\usepackage{xstring}
\usepackage{axessibility}  % Improves PDF readability for those with disabilities.

\usepackage[pagebackref=true,breaklinks=true,letterpaper=true,colorlinks,bookmarks=false]{hyperref}

\newcommand{\tabincell}[2]{\begin{tabular}{@{}#1@{}}#2\end{tabular}}
\newcommand{\x}{\boldsymbol{x}}

\iccvfinalcopy % *** Uncomment this line for the final submission

\def\iccvPaperID{7791} % *** Enter the ICCV Paper ID here

\ificcvfinal\fi

\begin{document}

%%%%%%%%% TITLE
\title{Semantic Concentration for Domain Adaptation}

\author{
 Shuang Li\textsuperscript{1} \space
 Mixue Xie\textsuperscript{1} \space
 Fangrui Lv\textsuperscript{1} \space
 Chi Harold Liu\textsuperscript{1\thanks{Corresponding author.}} \space\space
 Jian Liang\textsuperscript{\rm2} \space
 Chen Qin\textsuperscript{\rm3} \space
 Wei Li\textsuperscript{4} \space \vspace{.3em}\\
 \textsuperscript{1}Beijing Institute of Technology\quad\textsuperscript{2}Alibaba Group\quad\textsuperscript{3}University of Edinburgh\quad\textsuperscript{4}Inceptio Tech.\\
 {\tt\small shuangli@bit.edu.cn \space michellexie102@gmail.com \space fangruilv@bit.edu.cn \space liuchi02@gmail.com} \\
 {\tt\small liangjianzb12@gmail.com \space Chen.Qin@ed.ac.uk \space liweimcc@gmail.com}\\
}

\maketitle
\ificcvfinal\thispagestyle{empty}\fi

\begin{abstract}
   Domain adaptation (DA) paves the way for label annotation and dataset bias issues by the knowledge transfer from a label-rich source domain to a related but unlabeled target domain. A mainstream of DA methods is to align the feature distributions of the two domains. However, the majority of them focus on the entire image features where irrelevant semantic information, e.g., the messy background, is inevitably embedded. Enforcing feature alignments in such case will negatively influence the correct matching of objects and consequently lead to the semantically negative transfer due to the confusion of irrelevant semantics. To tackle this issue, we propose Semantic Concentration for Domain Adaptation (SCDA), which encourages the model to concentrate on the most principal features via the pair-wise adversarial alignment of prediction distributions. Specifically, we train the classifier to class-wisely maximize the prediction distribution divergence of each sample pair, which enables the model to find the region with large differences among the same class of samples. Meanwhile, the feature extractor attempts to minimize that discrepancy, which suppresses the features of dissimilar regions among the same class of samples and accentuates the features of principal parts. As a general method, SCDA can be easily integrated into various DA methods as a regularizer to further boost their performance. Extensive experiments on the cross-domain benchmarks show the efficacy of SCDA.
\end{abstract}

%%%%%%%%% BODY TEXT

\section{Introduction}

\begin{figure}[htbp]
    \centering
    \includegraphics[width=0.39\textwidth]{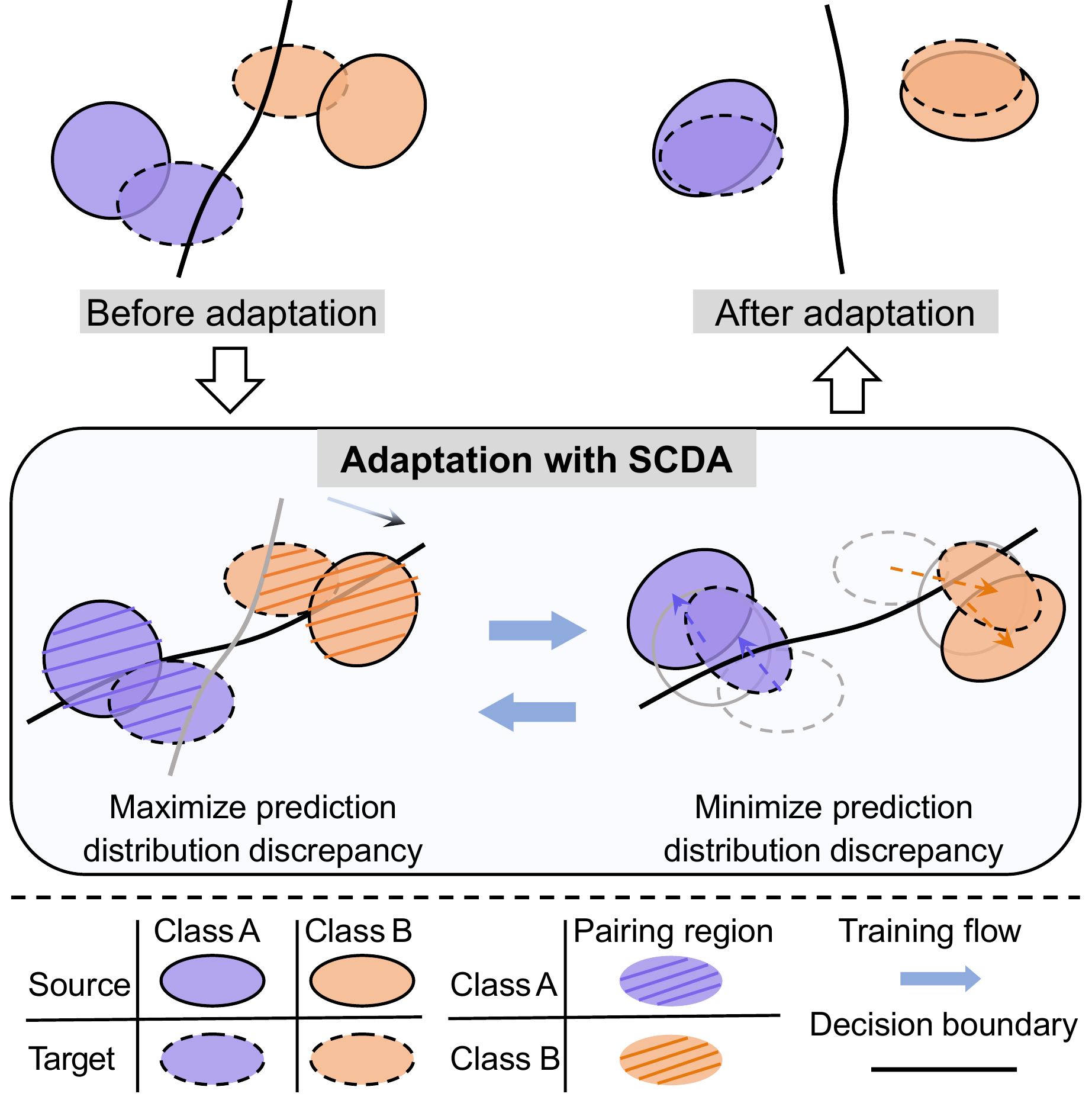}
    \caption{Illustration of the adversarial process of SCDA at the macro level. Classifier is trained to maximize the prediction distribution discrepancy of samples in the pairing region, which causes the decision boundary to pass through the high density area of the pairing region. While the feature extractor tries to minimize that discrepancy, which pushes the features away from the decision boundary. Finally, well-aligned features can be obtained through the adversarial game between the classifier and feature extractor.}
    \vspace{-4mm}
    \label{Fig_Intro}
\end{figure}

Deep neural network (DNN) has achieved great success in diverse machine learning problems \cite{alexnet,chen2018deeplab,ren2015faster}. Unfortunately, the impressive performance gain heavily relies on the access to massive well-labeled training data. And it is often time and cost prohibitive to manually annotate sufficient training data in practice. Besides, another drawback of conventional deep learning is the poor generalization on a new dataset, due to the domain shift issue \cite{A-distance,survey,ben2007analysis}. Hence, there is a strong motivation to utilize the knowledge of a label-rich domain (i.e., source domain) to assist the learning in a related but unlabeled domain (i.e., target domain), which is often referred to as domain adaptation (DA).

To alleviate the domain shift problem, the common practice of DA is to reduce the cross-domain distribution discrepancy by learning domain-invariant feature representations. Generally, these DA methods can be roughly categorized as the discrepancy-based methods \cite{DAN,JAN,DeCAF,MMD}, which align the domain distributions by minimizing a well-designed statistic metric, and the adversarial-based methods \cite{DANN,CDAN,ADDA,MCD,JADA,MADA}, where the domain discriminator is designed to distinguish between source and target samples and the feature extractor tries to confuse the discriminator.

Although these DA methods have admittedly achieved promising results, most of them use the features encoded without emphasis to match the feature distributions of two domains. In such case, irrelevant semantic information, e.g., the messy background is inevitably embedded, which may negatively influence the correct matching of objects and consequently lead to the semantically negative transfer. 

To relieve this issue, we propose to achieve the Semantic Concentration for Domain Adaptation (SCDA) by leveraging the dark knowledge \cite{cskd} (i.e., knowledge on the wrong predictions). Actually, SCDA is motivated by the findings in \cite{CAM} that the class prediction made by the model depends on what it has concentrated on and the concentrated region for each class prediction can be located with the feature maps and corresponding classification weights. Thus, we expect to find the concentrated regions for wrong predictions and suppress the features of these regions when encoding the image into features.

For this purpose, we propose to class-wisely align the pair-wise prediction distributions in an adversarial manner, which is shown in Fig. \ref{Fig_Intro}. Samples of the same label from two domains compose the pairing region for each class. The paring of samples includes intra-domain paring (i.e., pairing within source domain) and inter-domain pairing (i.e., pairing between source and target samples). For any sample pair of the same label, the classifier is trained to maximize their prediction distribution discrepancy, while the feature extractor strives to minimize that discrepancy. From the micro perspective, when the feature extractor is fixed, maximizing the prediction distribution discrepancy of the sample pair will cause the classification weights for dark knowledge to be larger. Then to reduce that discrepancy, features of these dark knowledge have to be suppressed, since the classification weights for them became larger in the previous training of the classifier. From the macro perspective, to maximize the prediction discrepancy in the pairing region with the feature extractor fixed, the decision boundary will cross the high density area of the pairing region. Then, to reduce the discrepancy, features will be pushed away from the decision boundary. Finally, the model is able to concentrate on the most principal features and achieves well-aligned features class-wisely via the min-max game.

Our contributions are summarized as follows:
\begin{itemize}
      \item This paper proposes a novel adversarial method for DA, i.e., the pair-wise adversarial alignment of prediction distribution discrepancy. Our method can suppress the irrelevant semantic information and accentuate the class object when encoding features, thus achieving the semantic concentration.
      \item As a simple and generic method, SCDA can be easily integrated as a regularizer into various DA methods and greatly improve their adaptation performances.
      \item Extensive experimental results and analysis demonstrate that SCDA greatly suppresses irrelevant semantics during the adaptation process, yielding state-of-the-art results on multiple cross-domain benchmarks.
\end{itemize}
\section{Related Work}

\textbf{Feature Distribution Alignment.} The distribution discrepancy between domains poses a great challenge for domain adaptation. To address this issue, the existing DA methods can be roughly divided into two categories. One is the statistical discrepancy based methods which aim to match various statistical moments across domains~\cite{RTN, JAN, MDD, DeepCORAL, ETD}. For instance, MDD \cite{MDD} introduces the margin disparity discrepancy to reduce the distribution discrepancy with a rigorous generalization bound. And based on the Earth Mover's distance, \cite{ETD} proposes an enhanced transport distance (ETD) to minimize the feature alignment loss. %Further, \cite{GSP} leverages the Gromov-Wasserstein (GW) discrepancy \cite{GW} to achieve the edge-level feature alignment.

The other category is inspired by the generative adversarial network (GAN) \cite{GAN}, which aims to learn domain-invariant features by playing a two-player min-max game \cite{DANN, CDAN, JADA, ADDA, MCD, GVB, SymNets}. For example, DANN \cite{DANN} and CDAN \cite{CDAN} introduce a domain discriminator to play the min-max game where the domain discriminator strives to distinguish source samples from target samples while the feature extractor tries to confuse the domain discriminator.

However, these methods focus on the alignment of the entire image features. The irrelevant semantic information e.g., messy backgrounds, may predominate the adaptation process, leading to samples of different categories misaligned or samples in the same category unaligned.

\textbf{Concentration Mechanism.} There have been recent efforts toward boosting the adaptation performance via applying different degrees of concentration on distinct image regions \cite{RMVA, Attention}. Several approaches leverage the attention-based methods to weight features at the pixel level, which facilitates the model concentrating on and transferring more principal semantic information across domains. \cite{DUCDA, DAAA, TADA} utilize attention mechanism to transfer features with high correlations across two distributions. DUCDA \cite{DUCDA} develops an attention transfer mechanism for DA, which transfers the knowledge of discriminative patterns of source images to target. Differently, instead of exploring the space attention knowledge, DCAN \cite{DCAN} explores the low-level domain-dependent knowledge in the channel attention. 

Although these attention-based DA methods can also suppress features of irrelevant semantics, most of them need to elaborately design a complex network architecture to derive the appropriate concentrations, greatly limiting their versatility. By contrast, our method leverages the pair-wise adversarial alignment on prediction space to achieve the concentration, which is easy to implement and can be used as a plug-and-play regularizer to various DA methods to further boost their performance.

\textbf{Dark Knowledge.} Numerous DA methods have explored the prediction space to boost the feature generation \cite{MCD,CDAN,RCA}, while most of them only focus on the correct class prediction. To fully leverage the prediction information, we introduce the concept of dark knowledge \cite{Distilling}, i.e., the knowledge on wrong predictions made by DNNs. In fact, dark knowledge is firstly proposed in knowledge distillation \cite{Distilling}, where the knowledge is transferred from a powerful teacher model to a student \cite{cskd,Zhang_2019_ICCV,2019Data}. For DA, the dark knowledge is also leveraged by some methods \cite{MCC,BCDM} to excavate information contained in non-target labels. MCC \cite{MCC} exploits the dark knowledge to formulate the tendency that a classifier confuses the predictions between the correct and ambiguous classes, and then minimizes the confusion. BCDM \cite{BCDM} proposes a novel metric using the dark knowledge of bi-classifiers to measure their discrepancy, where the classifiers are forced to produce more consistent predictions in a class-wise manner.

In this paper, we directly leverage the correspondence between the dark knowledge and its activated feature regions. By suppressing these features of dark knowledge via our proposed pair-wise adversarial alignment of predictions, we can effectively avoid the negative effect caused by irrelevant semantics in the adaptation process.
\section{Method}

\begin{figure*}[htbp]
    \centering
    \includegraphics[width=0.82\textwidth]{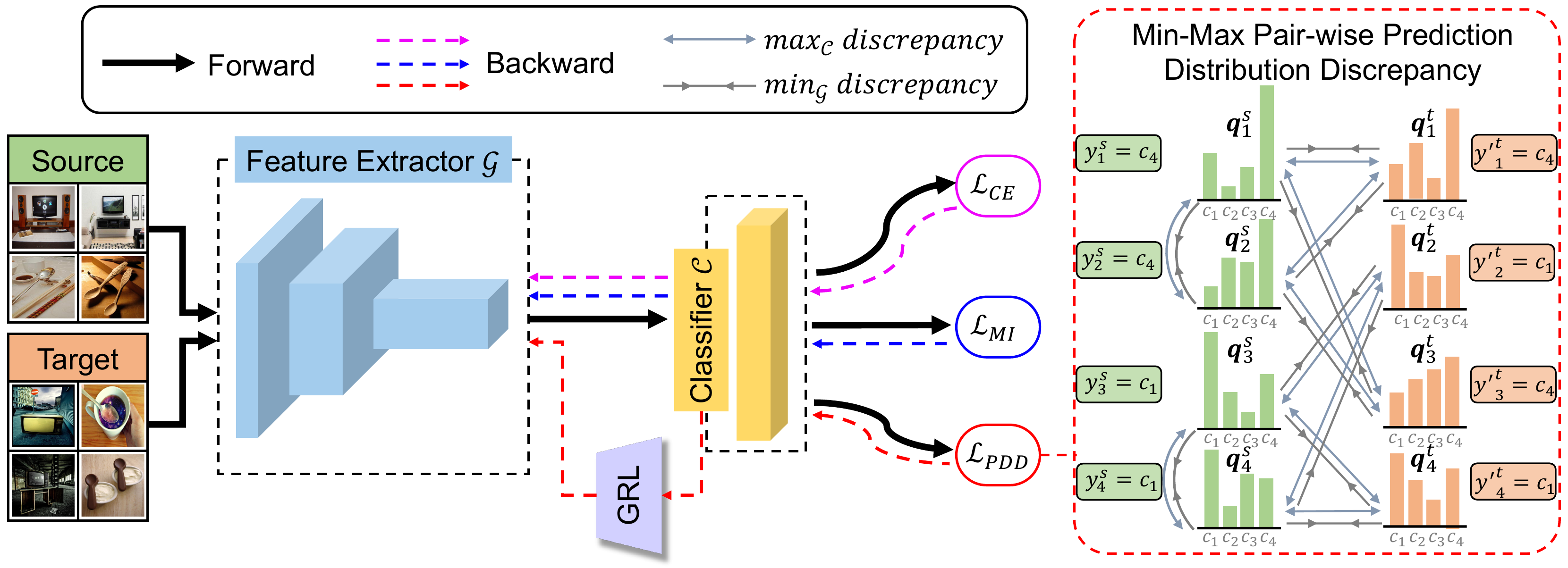}
    \caption{Overview of SCDA. $\{\boldsymbol{q}^{s}_i\}_{i=1}^{4}$ and $\{\boldsymbol{q}^{t}_j\}_{j=1}^{4}$ are the soften softmax predictions of a batch of source and target samples, respectively. GRL is the gradient reverse layer. $\mathcal{L}_{CE}$ is the cross-entropy loss on source domain. $\mathcal{L}_{MI}$ is the mutual information maximization loss on target domain. $\mathcal{L}_{PDD}$ is the pair-wise adversarial alignment loss of prediction distributions. The pairing of samples is shown in the right of the figure. The classifier is trained to maximize the prediction distribution discrepancy of each sample pair, while the feature extractor tries to minimize that discrepancy. Note that we use ground-truth labels for source samples, while pseudo labels for unlabeled target samples.}
    \vspace{-3mm}
    \label{Fig_method}
\end{figure*}

\subsection{Preliminaries and Motivation}

In DA, there are two domains accessible: a labeled source domain with $N_s$ samples, denoted as $\mathcal{S}=\{(\x^s_i, y^s_i)\}_{i=1}^{N_s}$ where $y^s_i \in \{1, 2, ..., C\}$ is the corresponding label of source sample $\x^s_i$, and an unlabeled target domain with $N_t$ samples, denoted as $\mathcal{T}=\{\x^t_j\}_{j=1}^{N_t}$. Source and target domains share the same label space, but differ in the data probability distributions. Such distribution discrepancy often leads to the performance degradation when the network trained on source domain is directly applied to target domain. In this paper, we denote the network by $\mathcal{F}$ which is composed of a feature extractor $\mathcal{G}$ and a classifier $\mathcal{C}$. The goal of DA is to adapt the network $\mathcal{F}$ from source to target by fully exploring the knowledge of labeled source data and unlabeled target data during the training procedure.

Most DA methods are based on the feature distribution alignment where entire image features are considered. But irrelevant semantics, e.g., messy backgrounds may also be embedded into the entire features and thus the predictions for the wrong classes may be relatively high without suppression for these features, which may result in the semantically negative transfer. Hence, it is necessary to find these concentrated regions for dark knowledge and suppress the features of these regions. Motivated by the close relationship among the prediction, classification weights and the features shown in \cite{CAM}, we propose Semantically Concentration for Domain Adaptation (SCDA), which leverages the pair-wise adversarial alignment of prediction distribution to suppress the features of dark knowledge and thus accentuates the features of principal parts for correct class. Briefly, we take the classifier and the feature extractor as the two players in the adversarial game. The classifier tries to increase the classification weights for wrong classes by maximizing the pair-wise prediction distribution discrepancy. While the feature extractor strives to suppress the features for the wrong classes to reduce that discrepancy. Via the min-max game, we can suppress the influence of irrelevant semantics on the feature alignment of two domains.

\subsection{Revisit the Class Activation Map}

In this section, we revisit the class activation map in \cite{CAM} to show the close relationship among the prediction, classification weights and features. For a particular class, its corresponding class activation map reflects which image region the model has concentrated on to make its prediction.

For a given image, let $a_h (u, v)$ denote the activation at spatial location $(u, v)$ of the $h$-th channel of the feature maps in the last convolutional layer. Then performing the global average pooling (GAP) on the $h$-th channel, we obtain $f_h$, i.e., $ f_h = \frac{1}{H W}\sum_{u,v} a_h (u, v)$, where $H$ and $W$ are the height and width of the feature map. For class $c$, the logit score $z_c$ given by the model is $\sum_h w^c_h f_h$, where $w^c_h$ is the classification weight (essentially the importance) of $h$-th feature map for class $c$. Here we omit the bias term, since it has no impact on the classification performance. Finally, the softmax score for class $c$ is $p_c = \frac{\exp(z_c)} {\sum_c \exp(z_c)}$.

Plugging $f_h = \frac{1}{H W}\sum_{u,v} a_h (u, v)$ into the expression of $z_c$, we can obtain
\begin{small}\begin{align}
    z_c & = \sum_h w^c_h \frac{1}{H W} \sum_{u,v} a_h(u,v) \notag \\
    & = \frac{1}{H W} \sum_{u,v} \sum_h w^c_h a_h(u,v) \notag \\
    & = \frac{1}{H W} \sum_{u,v} A_c(u,v),
\end{align}\end{small}where $A_c(u,v) = \sum_h w^c_h a_h(u,v)$. For a given model, $H W$ is a constant. Thus, $A_c(u,v)$ directly reflects the importance of the activation at the spatial location $(u, v)$ of the class activation map $A_c$ when classifying an image to class $c$. Finally, by upsampling the class activation map to the size of the original image, we can locate the regions concentrated on by the model for a particular class.

From the expressions of $z_c = \frac{1}{H W} \sum_{u,v} A_c(u,v)$ and $p_c = \frac{\exp(z_c)} {\sum_c \exp(z_c)}$, we can see that the prediction distribution of an image depends on the class activation maps, while the class activation maps reflect what the model has concentrated on. This motivates us to leverage the class activation maps of wrong predictions to find the regions that the model should not concentrate on and then suppress the features of these regions. Below we will describe how to achieve this idea via the pair-wise adversarial alignment of prediction distributions, which is the main component of our work.

\subsection{Amplify Concentrations on Irrelevant Regions}

Firstly, we describe the construction of our sample pairs, which is shown in Fig. \ref{Fig_method}. Samples of the same label from two domains compose the paring region for corresponding class. Since target domain is unlabeled, we employ the pseudo label predicted by the model for each target sample, i.e., ${y^{\prime}}^t_j = \mathop{\arg\max}_{c} {p^t_j}^{(c)}$ where ${p^t_j}^{(c)}$ is the $c$-th element of the softmax outputs of target sample $\x^t_j$. Two samples are considered as a pair if their labels are same. For each class, there exist two kinds of sample pairs, i.e., intra-domain sample pairs \footnote{Here, we do not conduct intra-domain pairing within target domain, since target data have no ground-truth labels.} (pairing within source domain) and inter-domain sample pairs (pairing between source and target domains).

To amplify the concentrations on irrelevant regions, we train the classifier to maximize the prediction distribution discrepancy of each sample pair. Since we have two kinds of sample pairs, the total loss of prediction distribution discrepancy includes the intra-domain and inter-domain parts, i.e., $\mathcal{L}_{PDD_{s,s}}$ and $\mathcal{L}_{PDD_{s,t}}$, which are denoted as
\begin{small}\begin{align}
        \max_{\mathcal{C}} & \; {\mathcal{L}}_{PDD_{s,s}} + {\mathcal{L}}_{PDD_{s,t}} \notag \\
        & = \frac{1}{M_{s,s}} T^2 \sum_{y^s_i = y^s_k} JS(\boldsymbol{q}^s_i, \boldsymbol{q}^s_k) \notag \\
        & + \frac{1}{M_{s,t}} T^2 \sum_{y^s_i = {y^{\prime}}^t_j} JS(\boldsymbol{q}^s_i, \boldsymbol{q}^t_j).
        \label{Eq:maximize_discrepancy}
\end{align}\end{small}Here, we use \textit{Jensen–Shannon} (JS) divergence to measure the discrepancy between a pair of predictions, due to its symmetry and finiteness compared with \textit{Kullback-Leibler} divergence. $\boldsymbol{q}^s_i = \mathrm{softmax} (\mathcal{F}(\x^s_i) / T)$, where $T$ is the temperature scaling parameter. To avoid the gradient vanishing, we multiply $T^2$ to maintain the magnitudes of gradients. $M_{s,s}$ and $M_{s,t}$ represent the number of samples satisfying $y^s_i = y^s_k$ and $y^s_i = {y^{\prime}}^t_j$, respectively.

When the feature extractor is fixed, the class activation map only depends on the classification weights of the classifier. Since the sample pair belongs to the same class and the predictive scores for this class are both high, to maximize the prediction distribution discrepancy of the sample pair, the classification weights for other wrong classes increase. Thus, the irrelevant regions concentrated on by the model become more activated. Taking the ``Bike'' class for example, one image describes that a boy wearing a helmet is riding a bike and the other image describes a bike with flowers in the basket. For these two images, predictive scores for ``Bike'' are both high, such as with the prediction distributions of $[0.01, 0.79, 0.20]$ and $[0.15, 0.84, 0.01]$ respectively in the class order of ``Flowers'', ``Bike'' and ``Helmet''. The prediction distribution discrepancy mainly exists in the predictive scores for ``Flowers'' and ``Helmet''. To maximize the discrepancy, the former image will increase the predictive score for ``Helmet'', while the latter image will increase the score for ``Flowers'', which will cause the region of ``Helmet'' and ``Flowers'' with more concentration for the two images, respectively. By doing so, we amplify the concentrations on irrelevant regions.

\subsection{Suppress Features of Irrelevant Semantics}

In the previous section, we have found the regions that the model has concentrated on for the predictions of irrelevant classes. Now, we expect to suppress the features of these regions for a purer knowledge transfer in DA. To this end, we train the feature extractor to minimize the prediction distribution discrepancy of sample pairs, the loss of which is expressed as
\vspace{-3mm}
\begin{small}\begin{align}
        \min_{\mathcal{G}} & \; {\mathcal{L}}_{PDD_{s,s}} + {\mathcal{L}}_{PDD_{s,t}} \notag \\
        & = \frac{1}{M_{s,s}} T^2 \sum_{y^s_i = y^s_k} JS(\boldsymbol{q}^s_i, \boldsymbol{q}^s_k) \notag \\
        & + \frac{1}{M_{s,t}} T^2 \sum_{y^s_i = {y^{\prime}}^t_j} JS(\boldsymbol{q}^s_i, \boldsymbol{q}^t_j).
        \label{Eq:minimize_discrepancy}
\end{align}\end{small}Since the classification weights for wrong classes increased in the previous training of the classifier, to reduce the prediction distribution discrepancy, the feature extractor has to suppress the features of these irrelevant semantics and accentuate the features of similar parts in the sample pair. In the adversarial manner, for the intra-domain sample pairs, we can achieve the extraction of the most principal features for each class, which serves as good \textit{teachers} for target domain. For the inter-domain sample pairs, the negative influence of domain shift is reduced and more emphasis is laid on the transfer of common knowledge across two domains.

\subsection{Overall Formulation}

Different from previous work \cite{MCD, adversarial_dropout} that use alternate updating to achieve the adversarial manner, we leverage the gradient reverse layer (GRL) in Fig. \ref{Fig_method} to achieve the optimization of all network parameters with the stochastic gradient descent. The overall loss function is defined as
\begin{small}\begin{equation}
    {\mathcal{L}_{SCDA}} = {\mathcal{L}}_{CE} - \alpha {\mathcal{L}}_{PDD} - \beta {\mathcal{L}}_{MI},
\end{equation}\end{small}where $\alpha$ and $\beta$ are two positive trade-off parameters.

${\mathcal{L}}_{CE}$ is the standard cross-entropy loss to supervise the learning on source domain, which is denoted as
\begin{small}\begin{align}
    \min_{\mathcal{F}} {\mathcal{L}}_{CE} = \frac{1}{N_s} \sum_{i=1}^{N_s} \mathcal{E} (\mathcal{F}(\x^s_i), y^s_i),
    \label{Eq:loss_CE}
\end{align}\end{small}where $\mathcal{E} (\cdot, \cdot)$ is the cross-entropy loss function.

${\mathcal{L}}_{PDD}$ is our proposed adversarial loss of the prediction distribution discrepancy to achieve the semantic concentration for DA, the expression of which is denoted as
\begin{small}\begin{align}
    \min_{\mathcal{G}} \max_{\mathcal{C}}{\mathcal{L}}_{PDD} & = {\mathcal{L}}_{PDD_{s,s}} + {\mathcal{L}}_{PDD_{s,t}}.
    \label{Eq:PDD}
\end{align}\end{small}To avoid the tedious updating steps in alternate updating, we leverage the gradient reverse layer in \cite{DANN} to achieve the adversarial training by one back-propagation.

${\mathcal{L}}_{MI}$ is the mutual information maximization loss on target domain, which is introduced to improve the quality of pseudo labels. The expression of ${\mathcal{L}}_{MI}$ is
\begin{small}\begin{align}
        \max_{\mathcal{F}} {\mathcal{L}}_{MI} & = H(\hat{Y}) - H(\hat{Y} | X) \notag\\
        & = - \sum_{c=1}^{C} \hat{p}^{(c)} \log \hat{p}^{(c)} + \frac{1}{N_t} \sum_{j=1}^{N_t}  \langle \boldsymbol{p}^t_j , \log \boldsymbol{p}^t_j  \rangle,
        \label{Eq:loss_MI}
\end{align}\end{small}where $\boldsymbol{p}^t_j$ is the softmax prediction of target sample $\x^t_j$, $\hat{p}^{(c)}$ is the $c$-th element of $\hat{\boldsymbol{p}} = \frac{1}{N_t} \sum_{j=1}^{N_t} \boldsymbol{p}^t_j$ and $\langle \cdot, \cdot \rangle$ is the inner product operation. Actually, the second term of ${\mathcal{L}}_{MI}$ is equivalent to the entropy minimization \cite{entropy_minimization}, which is a generic technique used in DA methods to enhance the discriminability of the model for target data, e.g., \cite{SymNets, RTN, DIRT-T}. However, the entropy minimization may result into collapsed trivial solutions \cite{MEDM}. To avoid this, we introduce the first term of $\mathcal{L}_{MI}$ to ensure the diversity of predictions. Besides, we also set a threshold of 0.8 to select target samples with relatively correct classification, i.e., only $\{\x^t_j | \mathop{\max}_c {p^t_j}^{(c)}\ge0.8 \}$ participate in inter-domain pairing.

The effects of different loss terms will be analyzed in details in the ablation study.

\subsection{Regularizer to Existing DA Methods}

As a simple but powerful method, SCDA is orthogonal to most existing DA methods and can be easily integrated into them as a regularizer to bring remarkable improvements by simply adding a gradient reverse layer. Taking CDAN \cite{CDAN} as an example, the integrated loss is formulated as:
\begin{small}\begin{align}
    {\mathcal{L}}_{SCDA} + \gamma {\mathcal{L}}_{adv},
    \label{Eq:loss_general}
\end{align}\end{small}where $\gamma$ is the trade-off parameter and ${\mathcal{L}}_{adv}$ is the domain-adversarial loss for the domain discriminator in CDAN. We suggest that readers refer to \cite{CDAN} for the detailed formulation of $\mathcal{L}_{adv}$. The adversarial process in \cite{CDAN} is that domain discriminator strives to correctly classify the domain labels of samples while the feature extractor aims to generate features that can deceive the domain discriminator. In addition, our method can also be plugged into other DA methods, such as statistical discrepancy based methods \cite{MDD}. We will show the effects of SCDA as a regularizer in experiments.
\begin{table*}
  \setlength{\abovecaptionskip}{0.cm}
  \setlength{\belowcaptionskip}{0.cm}
  \caption{Accuracy(\%) on DomainNet for UDA (ResNet-101). In each sub-table, the column-wise domains are selected as the source domain and the row-wise domains are selected as the target domain. [$\dagger$ Implement according to source code.]}
 \centering
 \resizebox{\textwidth}{!}{
 \setlength{\tabcolsep}{0.5mm}{
   \begin{tabular}{|c|ccccccc||c|ccccccc||c|ccccccc|}
   \hline
   \textbf{ADDA} \cite{ADDA}  & clp   & inf   & pnt   & qdr   & rel   & skt   & Avg. & \textbf{DANN} \cite{DANN}  & clp   & inf   & pnt   & qdr   & rel   & skt   & Avg.  & \textbf{MIMTFL} \cite{MIMTFL}  & clp   & inf   & pnt   & qdr   & rel   & skt   & Avg.  \\
   \hline
   \hline
   clp   & -     & 11.2  & 24.1  & 3.2   & 41.9  & 30.7  & 22.2  & clp   & -    & 15.5 & 34.8 & 9.5  & 50.8 & 41.4 & 30.4   & clp   & -    & 15.1 & 35.6 & 10.7 & 51.5 & 43.1 & 31.2  \\
   inf   & 19.1  & -     & 16.4  & 3.2   & 26.9  & 14.6  & 16.0  & inf   & 31.8 & -    & 30.2 & 3.8  & 44.8 & 25.7 & 27.3   & inf   & 32.1 & -    & 31.0 & 2.9  & 48.5 & 31.0 & 29.1  \\
   pnt   & 31.2  & 9.5   & -     & 8.4   & 39.1  & 25.4  & 22.7  & pnt   & 39.6 & 15.1 & -    & 5.5  & 54.6 & 35.1 & 30.0   & pnt   & 40.1 & 14.7 & -    & 4.2  & 55.4 & 36.8 & 30.2  \\
   qdr   & 15.7  & 2.6   & 5.4   & -     & 9.9   & 11.9  & 9.1   & qdr   & 11.8 & 2.0  & 4.4  & -    & 9.8  & 8.4  & 7.3    & qdr   & 18.8 & 3.1  & 5.0  & -    & 16.0 & 13.8 & 11.3  \\
   rel   & 39.5  & 14.5  & 29.1  & 12.1  & -     & 25.7  & 24.2  & rel   & 47.5 & 17.9 & 47.0 & 6.3  & -    & 37.3 & 31.2   & rel   & 48.5 & 19.0 & 47.6 & 5.8  & -    & 39.4 & 32.1  \\
   skt   & 35.3  & 8.9   & 25.2  & 14.9  & 37.6  & -     & 25.4  & skt   & 47.9 & 13.9 & 34.5 & 10.4 & 46.8 & -    & 30.7   & skt   & 51.7 & 16.5 & 40.3 & 12.3 & 53.5 & -    & 34.9  \\
   Avg.  & 28.2  & 9.3   & 20.1  & 8.4   & 31.1  & 21.7  & 19.8  & Avg.  & 35.7 & 12.9 & 30.2 & 7.1  & 41.4 & 29.6 & 26.1   & Avg.  & 38.2 & 13.7 & 31.9 & 7.2  & 45.0 & 32.8 & 28.1  \\
   \hline
   \hline
   \textbf{ResNet-101} \cite{resnet} & clp   & inf   & pnt   & qdr   & rel   & skt   & Avg.  & \textbf{CDAN$^{\dagger}$} \cite{CDAN}  & clp   & inf   & pnt   & qdr   & rel   & skt   & Avg.  &  \textbf{MDD$^{\dagger}$} \cite{MDD} & clp   & inf   & pnt   & qdr   & rel   & skt   & Avg.  \\
   \hline
   \hline
   clp   &  -   & 19.3  & 37.5  & 11.1  & 52.2  & 41.0  & 32.2  & clp  &  -   & 20.4 & 36.6 & 9.0  & 50.7 & 42.3 & 31.8  & clp   & -    & 20.5 & 40.7 & 6.2  & 52.5 & 42.1 & 32.4 \\
   inf   & 30.2 & -     & 31.2  & 3.6   & 44.0  & 27.9  & 27.4  & inf  & 27.5 &  -   & 25.7 & 1.8  & 34.7 & 20.1 & 22.0  & inf   & 33.0 & -    & 33.8 & 2.6  & 46.2 & 24.5 & 28.0 \\
   pnt   & 39.6 & 18.7  & -     & 4.9   & 54.5  & 36.3  & 30.8  & pnt  & 42.6 & 20.0 &  -   & 2.5  & 55.6 & 38.5 & 31.8  & pnt   & 43.7 & 20.4 & -    & 2.8  & 51.2 & 41.7 & 32.0 \\
   qdr   & 7.0  & 0.9   & 1.4   & -     & 4.1   & 8.3   & 4.3   & qdr  & 21.0 & 4.5  & 8.1  &  -   & 14.3 & 15.7 & 12.7  & qdr   & 18.4 & 3.0  & 8.1  & -    & 12.9 & 11.8 & 10.8 \\
   rel   & 48.4 & 22.2  & 49.4  & 6.4   & -     & 38.8  & 33.0  & rel  & 51.9 & 23.3 & 50.4 & 5.4  &  -   & 41.4 & 34.5  & rel   & 52.8 & 21.6 & 47.8 & 4.2  & -    & 41.2 & 33.5 \\
   skt   & 46.9 & 15.4  & 37.0  & 10.9  & 47.0  & -     & 31.4  & skt  & 50.8 & 20.3 & 43.0 & 2.9  & 50.8 &  -   & 33.6  & skt   & 54.3 & 17.5 & 43.1 & 5.7  & 54.2 & -    & 35.0 \\
   Avg.  & 34.4 & 15.3  & 31.3  & 7.4   & 40.4  & 30.5  & 26.6  & Avg. & 38.8 & 17.7 & 32.8 & 4.3  & 41.2 & 31.6 & 27.7  & Avg.  & 40.4 & 16.6 & 34.7 & 4.3  & 43.4 & 32.3 & 28.6 \\
   \hline
   \hline
   \tabincell{c}{\textbf{SCDA}} & clp   & inf   & pnt   & qdr   & rel   & skt   & Avg.  & \tabincell{c}{\textbf{CDAN} \\ \textbf{+SCDA}}  & clp   & inf   & pnt   & qdr   & rel   & skt   & Avg.  & \tabincell{c}{\textbf{MDD} \\ \textbf{+SCDA}} & clp   & inf   & pnt   & qdr   & rel   & skt   & Avg. \\
   \hline
   \hline
   clp   &  -   & 18.6 & 39.3 & 5.1  & 55.0 & 44.1 & 32.4     & clp  &  -   & 19.5 & 40.4 & 10.3 & 56.7 & 46.0 & 34.6  & clp   & -    & 20.4 & 43.3 & 15.2 & 59.3 & 46.5 & 36.9 \\
   inf   & 29.6 &   -  & 34.0 & 1.4  & 46.3 & 25.4 & 27.3     & inf  & 35.6 &  -   & 36.7 & 4.5  & 50.3 & 29.9 & 31.4  & inf   & 32.7 & -    & 34.5 & 6.3  & 47.6 & 29.2 & 30.1 \\
   pnt   & 44.1 & 19.0 &   -  & 2.6  & 56.2 & 42.0 & 32.8     & pnt  & 45.6 & 20.0 &  -   & 4.2  & 56.8 & 41.9 & 33.7  & pnt   & 46.4 & 19.9 & -    & 8.1  & 58.8 & 42.9 & 35.2 \\
   qdr   & 30.0 & 4.9  & 15.0  &  -  & 25.4 & 19.8 & 19.0     & qdr  & 28.3 & 4.8  & 11.5 &  -   & 20.9 & 19.2 & 17.0  & qdr   & 31.1 & 6.6  & 18.0 & -    & 28.8 & 22.0 & 21.3 \\
   rel   & 54.0 & 22.5 & 51.9 & 2.3  &   -  & 42.5 & 34.6     & rel  & 55.5 & 22.8 & 53.7 & 3.2  &  -   & 42.1 & 35.5  & rel   & 55.5 & 23.7 & 52.9 & 9.5  & -    & 45.2 & 37.4 \\
   skt   & 55.6 & 18.5 & 44.7 & 6.4  & 53.2 &  -   & 35.7     & skt  & 58.4 & 21.1 & 47.8 & 10.6 & 56.5 &  -   & 38.9  & skt   & 55.8 & 20.1 & 46.5 & 15.0 & 56.7 & -    & 38.8 \\
   Avg.  & 42.6 & 16.7 & 37.0 & 3.6  & 47.2 & 34.8 & \textbf{30.3}    & Avg. & 44.7 & 17.6 & 38.0 & 6.6 & 48.2 & 35.8 & \textbf{31.8}  & Avg.  & 44.3 & 18.1 & 39.0 & 10.8  & 50.2 & 37.2 & \textbf{33.3} \\
   \hline
   \end{tabular}
   }}
 \label{tab:domainnet}
 \vspace{-1mm}
\end{table*}

% Table generated by Excel2LaTeX from sheet 'OFFICE-HOME'
\begin{table*}
    \caption{Accuracy (\%) on Office-Home for UDA (ResNet-50).}
    \footnotesize
    \begin{center}
    \setlength{\tabcolsep}{1.5mm}{
      \begin{tabular}{|l|cccccccccccc|c|}
      \hline
      Method & Ar$\rightarrow$Cl & Ar$\rightarrow$Pr & Ar$\rightarrow$Rw & Cl$\rightarrow$Ar & Cl$\rightarrow$Pr & Cl$\rightarrow$Rw & Pr$\rightarrow$Ar & Pr$\rightarrow$Cl & Pr$\rightarrow$Rw & Rw$\rightarrow$Ar & Rw$\rightarrow$Cl & Rw$\rightarrow$Pr & Avg \\
      \hline
      \hline
      ResNet-50 \cite{resnet}& 34.9 & 50.0 & 58.0 & 37.4 & 41.9 & 46.2 & 38.5 & 31.2 & 60.4 & 53.9 & 41.2 & 59.9 & 46.1 \\
      DANN \cite{DANN} & 45.6 & 59.3 & 70.1 & 47.0 & 58.5 & 60.9 & 46.1 & 43.7 & 68.5 & 63.2 & 51.8 & 76.8 & 57.6 \\
      JAN \cite{JAN} & 45.9 & 61.2 & 68.9 & 50.4 & 59.7 & 61.0 & 45.8 & 43.4 & 70.3 & 63.9 & 52.4 & 76.8 & 58.3 \\
      MCD \cite{MCD} & 48.9 & 68.3 & 74.6 & 61.3 & 67.6 & 68.8&  57.0 & 47.1 & 75.1 & 69.1 & 52.2 & 79.6 & 64.1 \\
      ETD \cite{ETD} & 51.3 & 71.9 & \textbf{85.7} & 57.6 & 69.2 & 73.7 & 57.8 & 51.2 & 79.3 & 70.2 & 57.5 & 82.1 & 67.3 \\
      SymNets \cite{SymNets} & 47.7 & 72.9 & 78.5 & 64.2 & 71.3 & 74.2 & 64.2 & 48.8 & 79.5 & 74.5 & 52.6 & 82.7 & 67.6 \\
      TADA \cite{TADA} & 53.1 & 72.3 & 77.2 & 59.1 & 71.2 & 72.1 & 59.7 & 53.1 & 78.4 & 72.4 & 60.0 & 82.9 & 67.6 \\
      GVB-GD \cite{GVB} & 57.0 & 74.7 & 79.8 & 64.6 & 74.1 & 74.6 & 65.2 & 55.1 & 81.0 & 74.6 & 59.7 & 84.3 & 70.4 \\
      \hline
      SCDA & 57.5 & 76.9 & 80.3 & 65.7 & 74.9 & 74.5 & 65.5 & 53.6 & 79.8 & 74.5 & 59.6 & 83.7 & 70.5 \\
      \hline
      CDAN \cite{CDAN} & 50.7 & 70.6 & 76.0 & 57.6 & 70.0 & 70.0 & 57.4 & 50.9 & 77.3 & 70.9 & 56.7 & 81.6 & 65.8 \\
      CDAN+SCDA & 57.1 & 75.9 & 79.9 & 66.2 & 76.7 & 75.2 & 65.3 & 55.6 & 81.9 & 74.7 & \textbf{62.6} & 84.5 & 71.3 \\
      \hline
      MDD \cite{MDD} & 54.9 & 73.7 & 77.8 & 60.0 & 71.4 & 71.8 & 61.2 & 53.6 & 78.1 & 72.5 & 60.2 & 82.3 & 68.1 \\
      MDD+SCDA & 58.9 & 77.2 & 81.0 & 66.6 & 75.5 & 75.9 & 64.1 & 56.3 & 82.2 & 73.3 & 61.5 & 84.3 & 71.4 \\
      \hline
      MCC \cite{MCC} & 55.1 & 75.2 & 79.5 & 63.3 & 73.2 & 75.8 & 66.1 & 52.1 & 76.9 & 73.8 & 58.4 & 83.6 & 69.4 \\
      MCC+SCDA & 57.1 & \textbf{79.1} & 82.7 & 67.7 & 75.3 & 77.6 & 66.3 & 52.5 & 81.9 & \textbf{74.9} & 60.1 & \textbf{85.0} & 71.7 \\
      \hline
      DCAN \cite{DCAN} & 54.5 & 75.7 & 81.2 & 67.4 & 74.0 & 76.3 & 67.4 & 52.7 & 80.6 & 74.1 & 59.1 & 83.5 & 70.5 \\
      DCAN+SCDA & \textbf{60.7} & 76.4 & 82.8 & \textbf{69.8} & \textbf{77.5} & \textbf{78.4} & \textbf{68.9} & \textbf{59.0} & \textbf{82.7} & \textbf{74.9} & 61.8 & 84.5 & \textbf{73.1} \\
      \hline
      \end{tabular}}
      \end{center}
      \label{tab:Office-Home}
      \vspace{-7mm}
\end{table*}

\section{Experiment}

\subsection{Experimental Setting}

\textbf{DomainNet} \cite{DomainNet} is the largest and the most challenging dataset for DA so far. It contains about 0.6 million images of 345 categories drawn from six diverse domains: Clipart (\textbf{clp}), Infograph (\textbf{inf}), Painting (\textbf{pnt}), Quickdraw (\textbf{qdr}), Real (\textbf{rel}) and Sketch (\textbf{skt}). Permuting the six domains, we build 30 adaptation tasks: \textbf{clp}$\rightarrow${\textbf{inf}}, ..., \textbf{skt}$\rightarrow${\textbf{rel}}.

\textbf{Office-Home} \cite{Office-Home} is a more challenging benchmark dataset for visual domain adaptation, which includes 15,500 images of 65 categories spreading in four distinct domains: Artistic images (\textbf{Ar}), Clip Art (\textbf{Cl}), Product images (\textbf{Pr}) and Real-World images (\textbf{Rw}). 12 adaptation tasks are constructed to evaluate our method, i.e., \textbf{Ar}$\rightarrow$\textbf{Cl}, ..., \textbf{Rw}$\rightarrow$\textbf{Pr}.

\textbf{Office-31} \cite{office-31} is a classical real-world benchmark dataset for DA. It contains 4,110 images of 31 classes shared by three distinct domains: Amazon (\textbf{A}), Webcam (\textbf{W}) and DSLR (\textbf{D}). We construct 6 adaptation tasks to evaluate our method, i.e., \textbf{A}$\rightarrow$\textbf{W}, ..., \textbf{D}$\rightarrow$\textbf{W}.

% Table generated by Excel2LaTeX from sheet 'OFFICE-31'
\begin{table}
  \begin{footnotesize}
  \centering
  \setlength{\abovecaptionskip}{0.cm}
  \setlength{\belowcaptionskip}{0.cm}
  \caption{Accuracy (\%) on Office-31 for UDA (ResNet-50). [Avg$^\ddagger$: mean values except D$\leftrightarrow$W]}
  \setlength{\tabcolsep}{0.5mm}{
  \begin{tabular}{|l|cccccc|c|c|}
  \hline
  Method & A$\rightarrow$W & D$\rightarrow$W & W$\rightarrow$D & A$\rightarrow$D & D$\rightarrow$A & W$\rightarrow$A & Avg & Avg$^\ddagger$ \\
  \hline
  \hline
  ResNet-50 \cite{resnet} & 68.4 & 96.7 & 99.3 & 68.9 & 62.5& 60.7 & 76.1 & 65.1 \\
  DANN \cite{DANN} & 82.0 & 96.9 & 99.1 & 79.7 & 68.2& 67.4 & 82.2 & 74.3 \\
  JAN \cite{JAN} & 85.4 & 97.4 & 99.8 & 84.7 & 68.6 & 70.0 & 84.3 & 77.2 \\
  CAT \cite{CAT} & 91.1 & 98.6 & 99.6 & 90.6 & 70.4 & 66.5 & 86.1 & 79.7 \\
  ETD \cite{ETD} & 92.1 & \textbf{100.0} & \textbf{100.0} & 88.0 & 71.0 & 67.8 & 86.2 & 79.7 \\
  MCD \cite{MCD} & 88.6 & 98.5 & \textbf{100.0} & 92.2 & 69.5 & 69.7 & 86.5 & 80.0 \\ 
  SymNets \cite{SymNets} & 90.8 & 98.8 & \textbf{100.0} & 93.9 & 74.6 & 72.5 & 88.4 & 83.0 \\
  TADA \cite{TADA} & 94.3 & 98.7 & 99.8 & 91.6 & 72.9 & 73.0 & 88.4 & 83.0 \\
  GVB-GD \cite{GVB} & 94.8 & 98.7 & \textbf{100.0} & 95.0 & 73.4 & 73.7 & 89.3 & 84.2 \\
  \hline
  SCDA & 94.2 & 98.7 & 99.8 & 95.2 & 75.7 & 76.2 & 90.0 & 85.3\\
  \hline
  CDAN \cite{CDAN} & 94.1 & 98.6 & \textbf{100.0} & 92.9 & 71.0 & 69.3 & 87.7 & 81.8 \\
  CDAN+SCDA & 94.7 & 98.7 & \textbf{100.0} & 95.4 & 77.1 & 76.0 & 90.3 & 85.8\\
  \hline
  MDD \cite{MDD} & 94.5 & 98.4 & \textbf{100.0} & 93.5 & 74.6 & 72.2 & 88.9 & 83.7  \\
  MDD+SCDA & 95.3 & 99.0 & \textbf{100.0} & 95.4 & 77.2 & 75.9 & \textbf{90.5} & \textbf{85.9} \\
  \hline
  MCC \cite{MCC} & \textbf{95.5} & 98.6 & \textbf{100.0} & 94.4 & 72.9 & 74.9 & 89.4 & 84.4 \\
  MCC+SCDA & 93.7 & 98.6 & \textbf{100.0} & \textbf{96.4} & 76.5 & 76.0 & 90.2 & 85.7 \\
  \hline
  DCAN \cite{DCAN} & 95.0 & 97.5 & \textbf{100.0} & 92.6 & 77.2 & 74.9 & 89.5 & 84.9 \\
  DCAN+SCDA & 94.8 & 98.2 & \textbf{100.0} & 94.6 & \textbf{77.5} & \textbf{76.4} & 90.3 & 85.8 \\
  \hline
  \end{tabular}}
  \label{tab:Office-31}
  \end{footnotesize}
  \vspace{-1mm}
\end{table}

\begin{table}[htbp]
  \footnotesize
  \centering
  \caption{Ablation Study of SCDA on Office-31 (ResNet-50).}
  \setlength{\tabcolsep}{0.3mm}{
  \begin{tabular}{|l|cccccc|c|}
    \hline
    Method & A$\rightarrow$W  & D$\rightarrow$W  & W$\rightarrow$D  & A$\rightarrow$D  & D$\rightarrow$A  & W$\rightarrow$A  & Avg \\
    \hline
    \hline
    ResNet-50 & 68.4 & 96.7 & 99.3 & 68.9 & 62.5 & 60.7 & 76.1 \\
    + SCDA (w/o $\mathcal{L}_{PDD}$) & 91.3 & 98.6 & 99.8 & 92.2 & 69.2 & 68.6 & 86.6 \\
    + SCDA (w/o $\mathcal{L}_{PDD_{s,t}}$) & 91.8 & 98.4 & 100.0 & 92.5 & 71.4 & 70.8 & 87.5 \\
    + SCDA (w/o $\mathcal{L}_{PDD_{s,s}}$) & 92.2 & 98.6 & 100.0 & 94.1 & 72.8 & 72.6& 88.3 \\
    + SCDA (w/o $\mathcal{L}_{MI}$) & 92.6 & 98.7 & 100.0 & 94.4 & 74.1 & 73.4& 88.9 \\
    + SCDA & 94.2 & 98.7 & 99.8 & 95.2 & 75.7 & 76.2 & \textbf{90.0} \\
    \hline
    \end{tabular}}
    \vspace{-4mm}
  \label{tab:ablation}
\end{table}% 

\textbf{Implementation details}. Following the standard protocol for DA \cite{DANN, CDAN, TAT}, we use all the labeled source data and unlabeled target data as training data and evaluate on unlabeled target data. We implement our approach in PyTorch framework\cite{paszke2019pytorch}. To fairly compare with existing methods, we use the same backbone networks, i.e., ResNet-50 \cite{resnet} pre-trained on ImageNet \cite{imagenet2014} for datasets: Office-31 and Office-Home, and ResNet-101 \cite{resnet} pre-trained on ImageNet \cite{imagenet2014} for DomainNet \cite{DomainNet}. In experiments, the input image size is cropped to 224 $\times$ 224. We employ the mini-batch stochastic gradient descent (SGD) optimizer with momentum of 0.9 and the learning rate strategy as described in \cite{DANN} for network optimization. To reduce the effect of unreliable predictions in the early training stage, we simply let the hyper-parameter $\alpha = \alpha_0 \rho$, where $\rho$ is the training progress changing from 0 to 1. And following MCC \cite{MCC}, we use Deep Embedded Validation (DEV) \cite{dev} to select the hyper-parameters and find that $T=10$, $\alpha_0=1.0$, $\beta=0.1$ works well on all datasets. Besides, the parameter sensitivity analysis is provided in section \ref{sec:analysis} to test the robustness of SCDA. Each adaptation task is evaluated by averaging the results of three random trials. Code of SCDA is available at \url{https://github.com/BIT-DA/SCDA.}

\subsection{Results}

\textbf{Results on DomainNet} are presented in Table~\ref{tab:domainnet}. Obviously, SCDA outperforms all the compared methods significantly in terms of the average accuracy. Particularly, applying SCDA to CDAN and MDD improves their prediction accuracy by 4.1\% and 4.7\% respectively. One interpretation is that our method suppresses the features of irrelevant semantics that may confuse the alignment process of CDAN and MDD. The encouraging results demonstrate the superiority of SCDA in processing complex datasets and its universality to existing DA methods.

\textbf{Results on Office-Home} are shown in Table~\ref{tab:Office-Home}, where we achieve comparable and even better performance, compared with these state-of-art DA methods. Moreover, our method achieves extra gain of 5.5\% and large improvements on \textbf{Cl} $\rightarrow$ \textbf{Ar}, \textbf{Cl}$\rightarrow$ \textbf{Pr}, \textbf{Cl}$\rightarrow$ \textbf{Rw} when applied to CDAN. The reason is that the images in \textbf{Cl} are rather complicated, while SCDA can purify the transferred knowledge by suppressing the features of irrelevant semantics. And DCAN+SCDA achieves the best performance of 73.1\%. These improvements validate the effectiveness of SCDA.

\textbf{Results on Office-31} are summarized in Table~\ref{tab:Office-31}. Obviously, we substantially obtain superior prediction accuracy over other popular adaptation methods. Particularly, when applying SCDA to MDD, we achieve the highest accuracy of 90.5\%. The outcomes show that SCDA is beneficial to promote the adaptation capability, especially on complex scenarios, e.g., \textbf{A} $\rightarrow$ \textbf{D}, \textbf{D} $\rightarrow$ \textbf{A} and \textbf{W} $\rightarrow$ \textbf{A}.

\begin{figure}[htbp]
  \centering
  \includegraphics[width=0.475\textwidth]{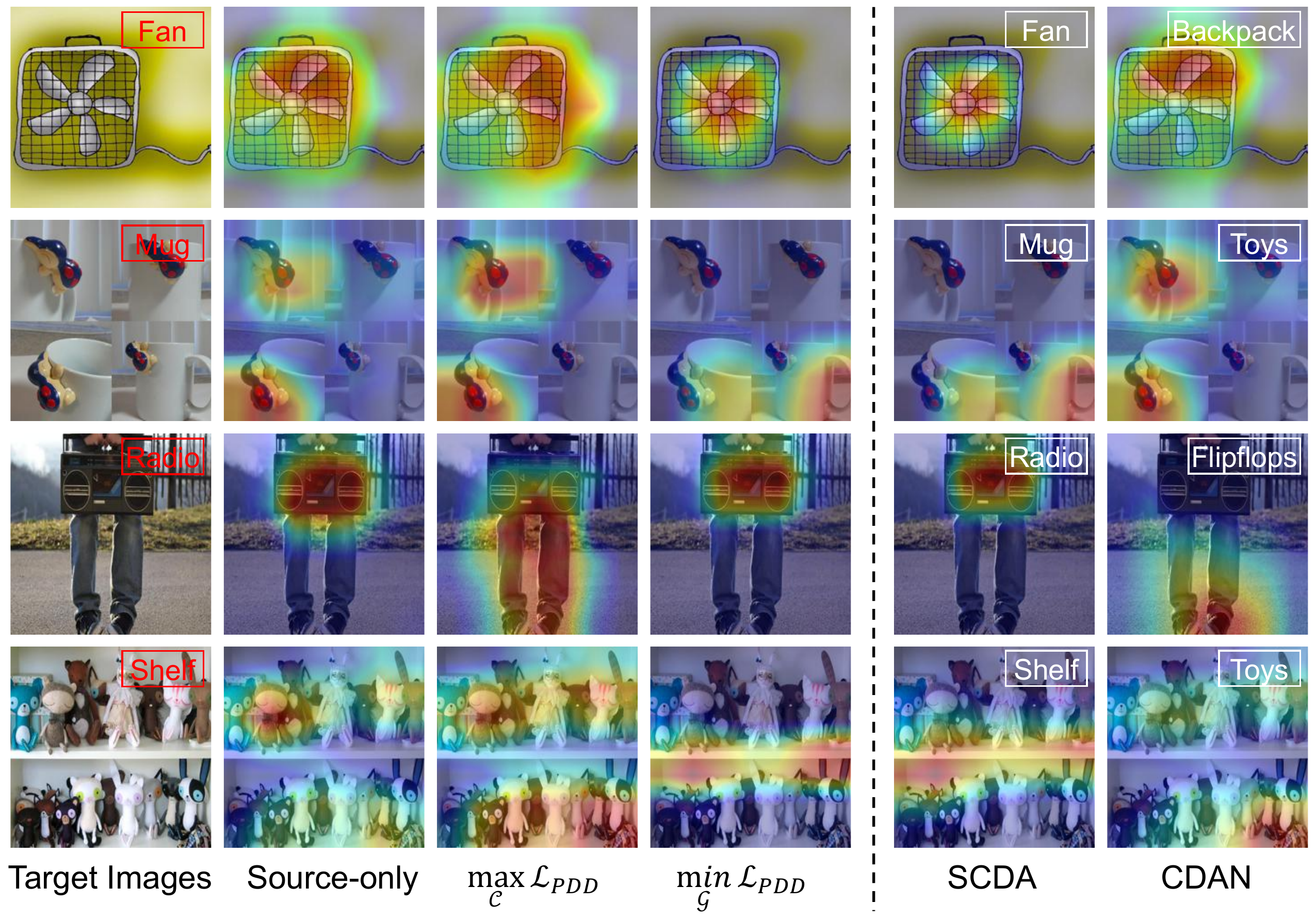}
  \caption{Concentration visualizations of the last convolutional layer of different methods on the task Rw $\rightarrow$ Ar of Office-Home. Here, the red font denotes the ground-truth labels, while the white font represents the pseudo labels predicted by different methods.}
  \vspace{-4mm}
  \label{Fig_Attention_Experiment}
\end{figure}

\subsection{Analysis}\label{sec:analysis}

\textbf{Ablation Study.} To investigate the efficacy of different components of SCDA, we conduct thorough ablation analysis on Office-31 based on ResNet-50: (1) SCDA (w/o $\mathcal{L}_{MI}$) denotes the variant of removing the mutual information maximization loss on target domain; (2) SCDA (w/o $\mathcal{L}_{PDD_{s,s}}$) and SCDA (w/o $\mathcal{L}_{PDD_{s,t}}$) respectively denote the variant of removing the pair-wise adversarial alignment of prediction distributions within source domain and cross domains; (3) SCDA (w/o $\mathcal{L}_{PDD}$) denotes the removal of both $\mathcal{L}_{PDD_{s,s}}$ and $\mathcal{L}_{PDD_{s,t}}$. The results are shown in Table \ref{tab:ablation}, where we can obviously see that full method SCDA outperforms other variants. While SCDA (w/o $\mathcal{L}_{PDD_{s,t}}$) suffers a obvious degradation of 2.5\%, which indicates the importance of transferring the common knowledge and suppressing domain-specific knowledge for DA problems by our loss $\mathcal{L}_{PDD_{s,t}}$. And SCDA is superior to SCDA (w/o $\mathcal{L}_{PPA_{s,s}}$), because $\mathcal{L}_{PPA_{s,s}}$ conduces to the constructing of good \textit{teachers} for target samples by learning the most principal features for classification. Besides, through improving the quality of pseudo labels for the paring process, SCDA achieves better performance than SCDA (w/o $\mathcal{L}_{MI}$).

\begin{figure*}[htbp]
  \centering
  \subfigure[W $\rightarrow$ A]{\includegraphics[width=0.215\textwidth]{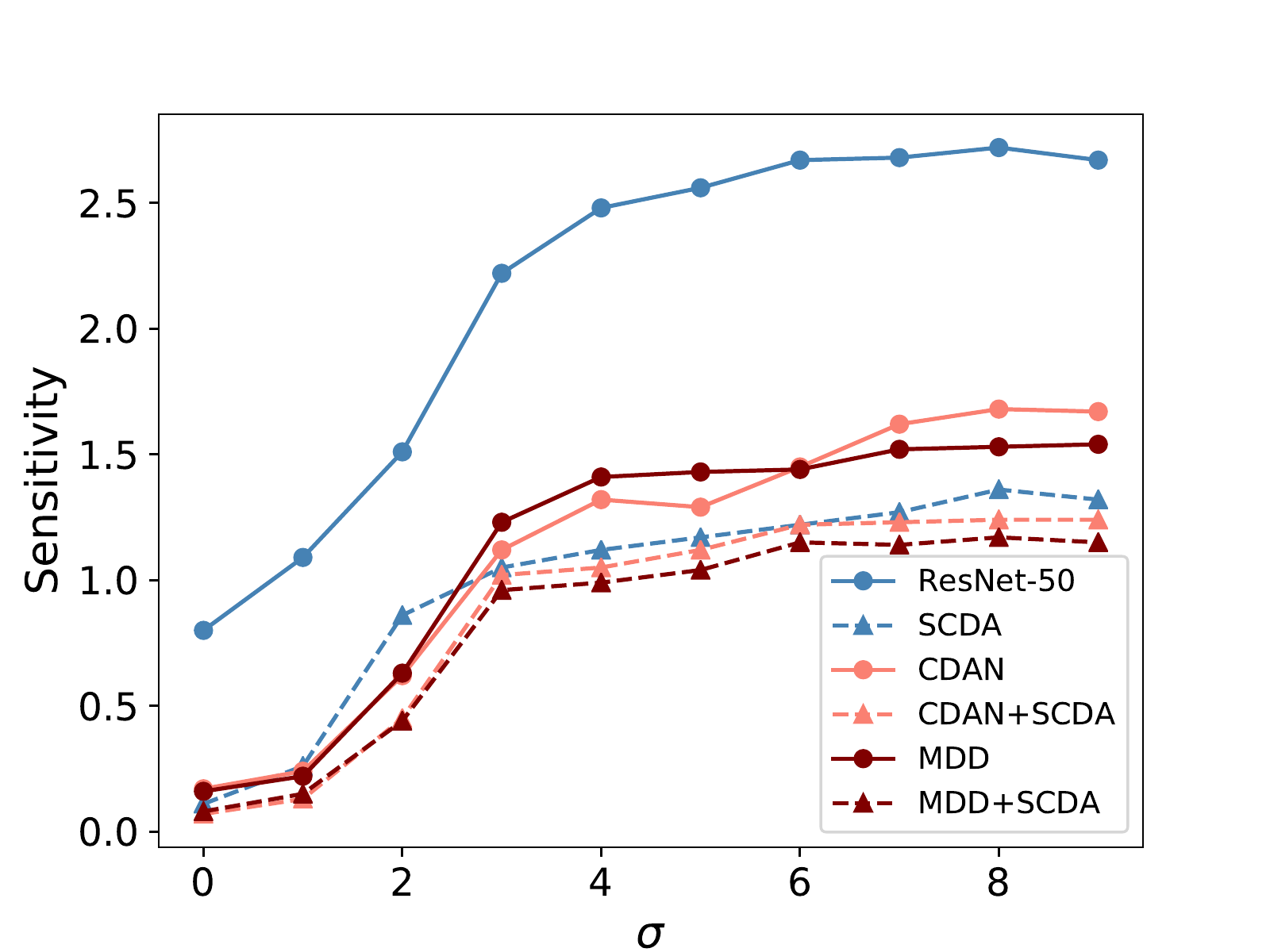} \label{Fig_Sensitivity}}
  \subfigure[A $\rightarrow$ D and A $\rightarrow$ W]{\includegraphics[width=0.215\textwidth]{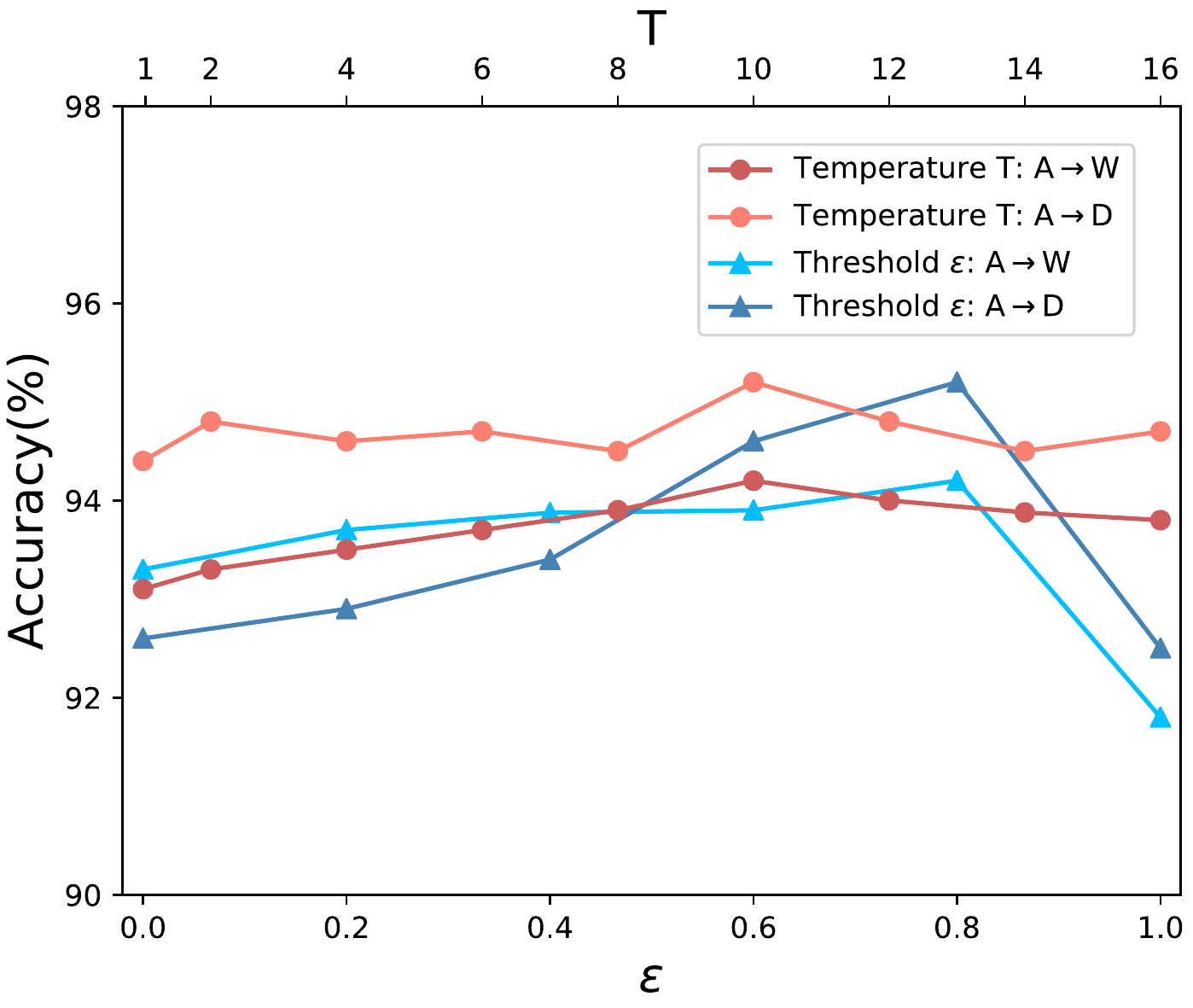} \label{Fig_ParameterSensitivity_curve}}
  \subfigure[A $\rightarrow$ D]{\includegraphics[width=0.22\textwidth]{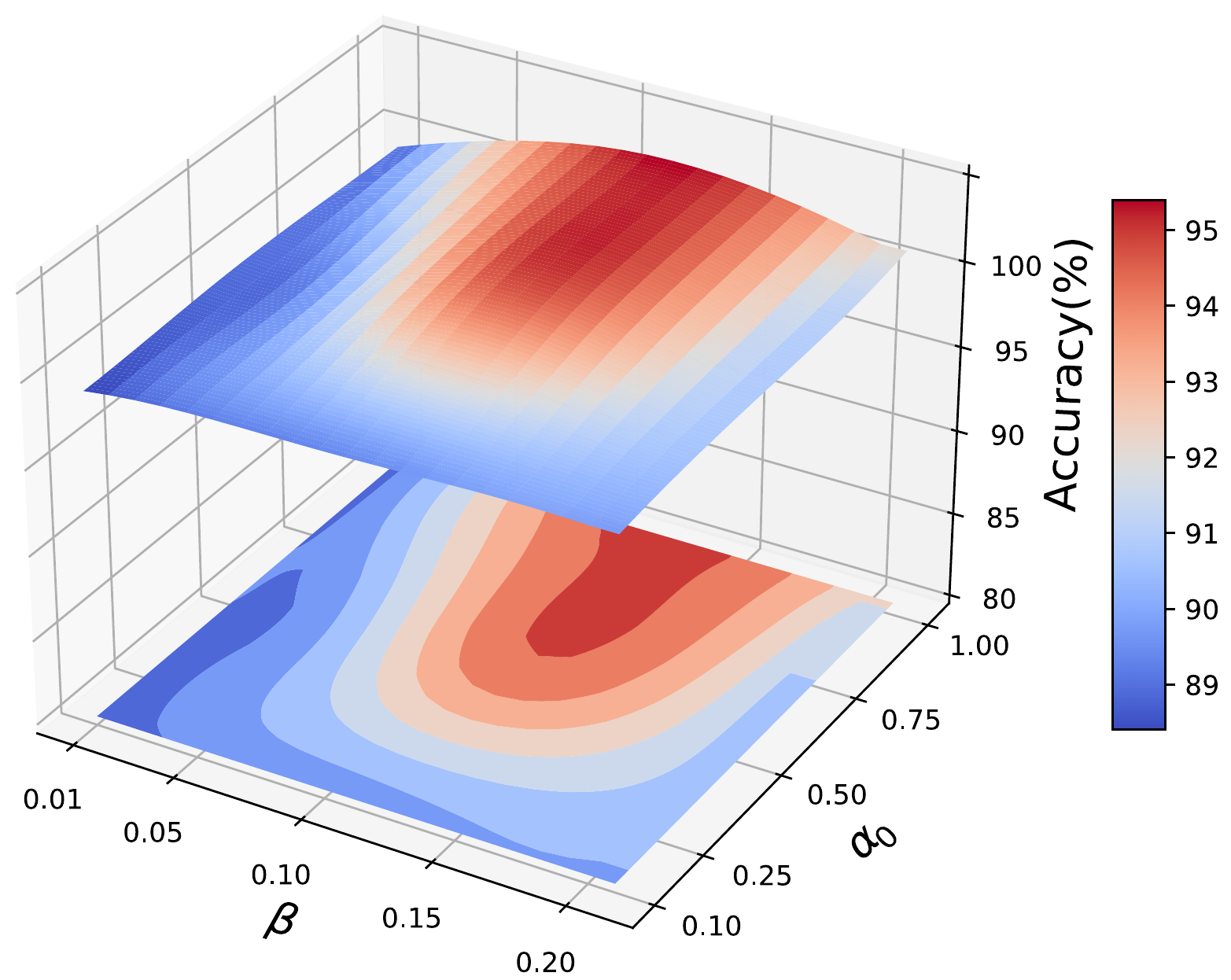} \label{Fig_ParameterSensitivity_A2D}}
  \subfigure[A $\rightarrow$ W]{\includegraphics[width=0.22\textwidth]{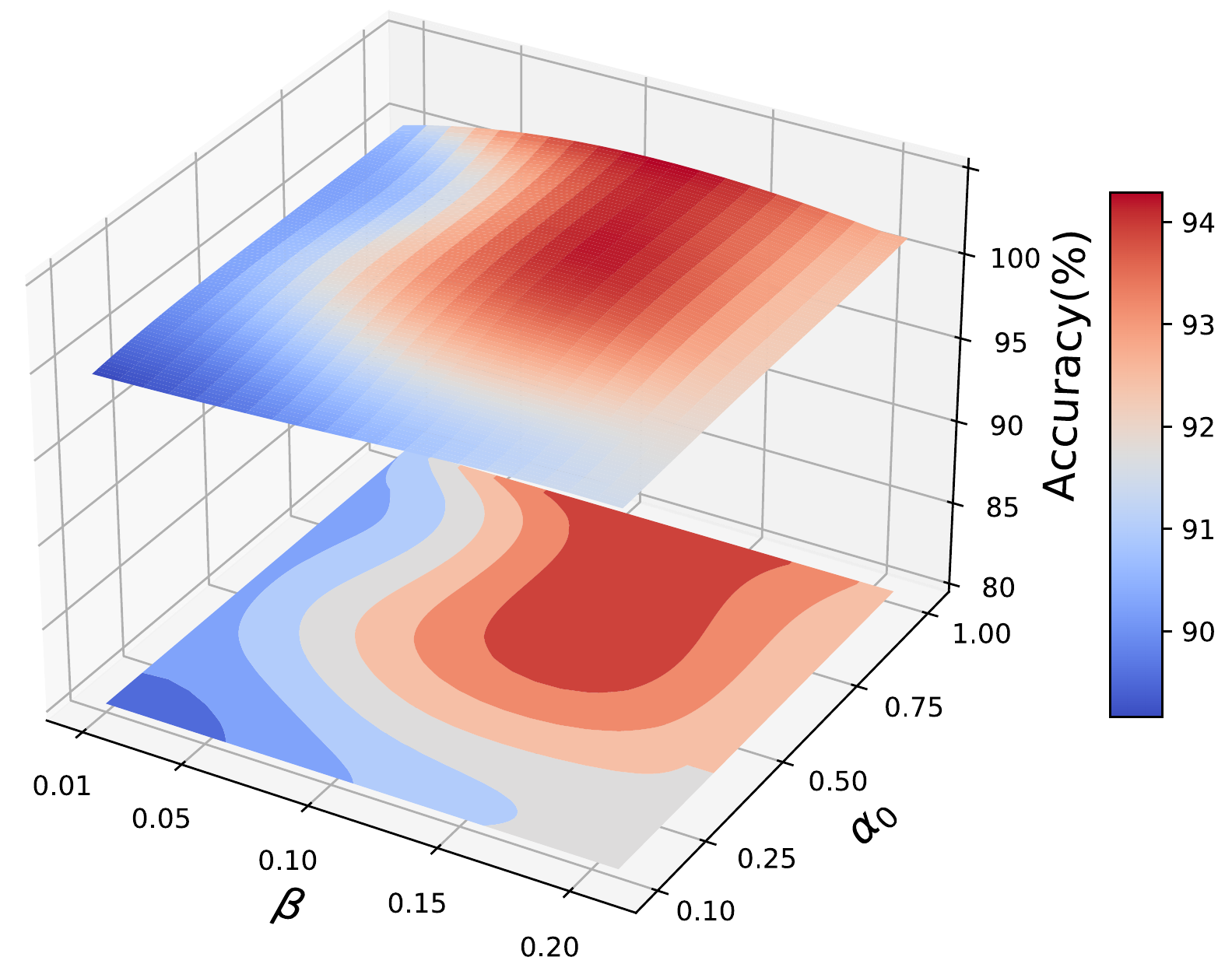} \label{Fig_ParameterSensitivity_A2W}}
  \caption{(a) is anti-jamming ability test of different methods on task W $\rightarrow$ A of Office-31 as the variance $\sigma$ of added Gaussian noise increasing from 0 to 10. (b) is the sensitivity of SCDA to parameters $T$ and $\epsilon$ on tasks A $\rightarrow$ D and A $\rightarrow$ W. (c) and (d) are the sensitivity of SCDA to parameters $\alpha_0$ and $\beta$ on tasks A $\rightarrow$ D and A $\rightarrow$ W, respectively.}
  \vspace{-3mm}
\end{figure*}

\begin{figure*}[htbp]
  \centering
  \subfigure[ResNet-50]{\includegraphics[width=0.235\textwidth]{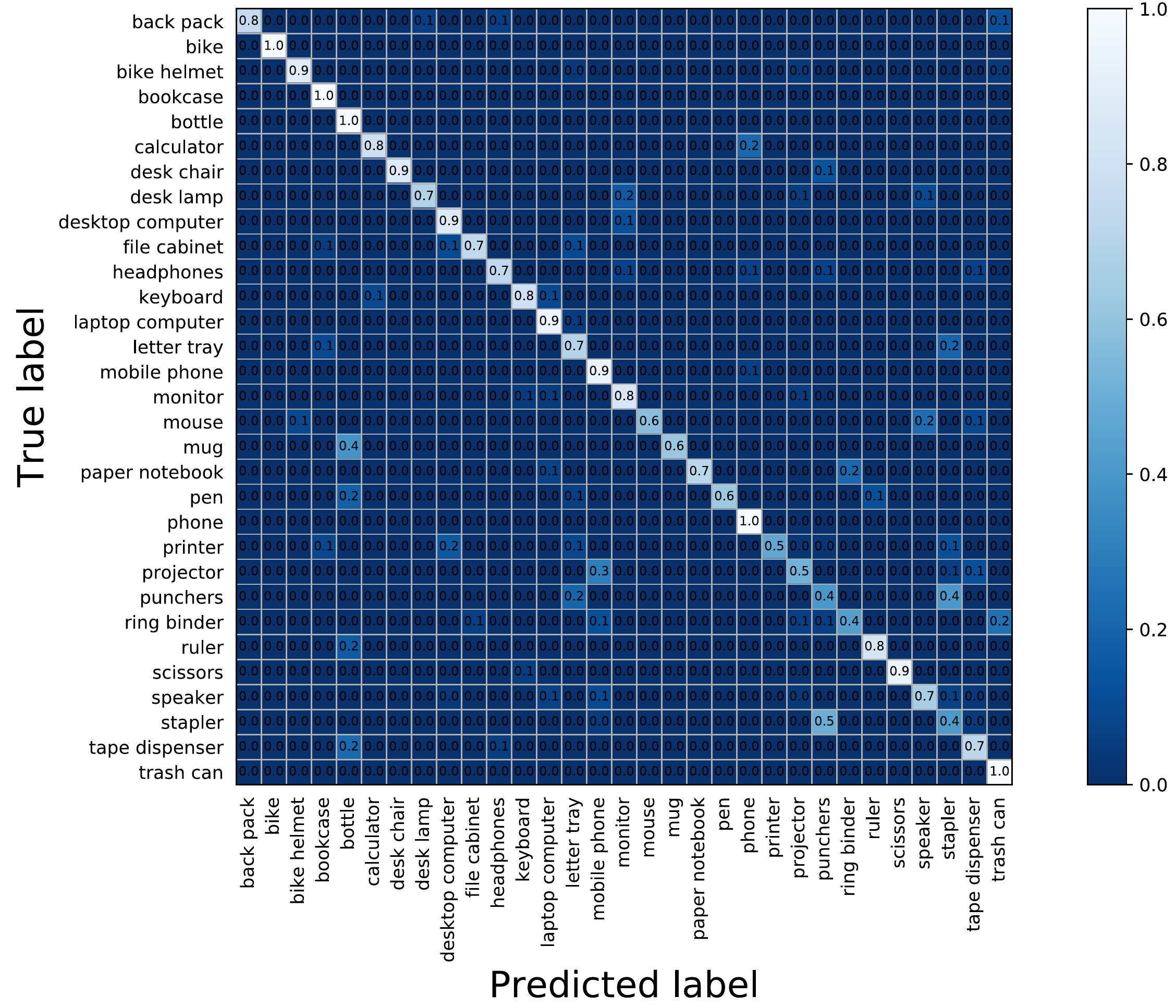}}
  \subfigure[SCDA]{\includegraphics[width=0.235\textwidth]{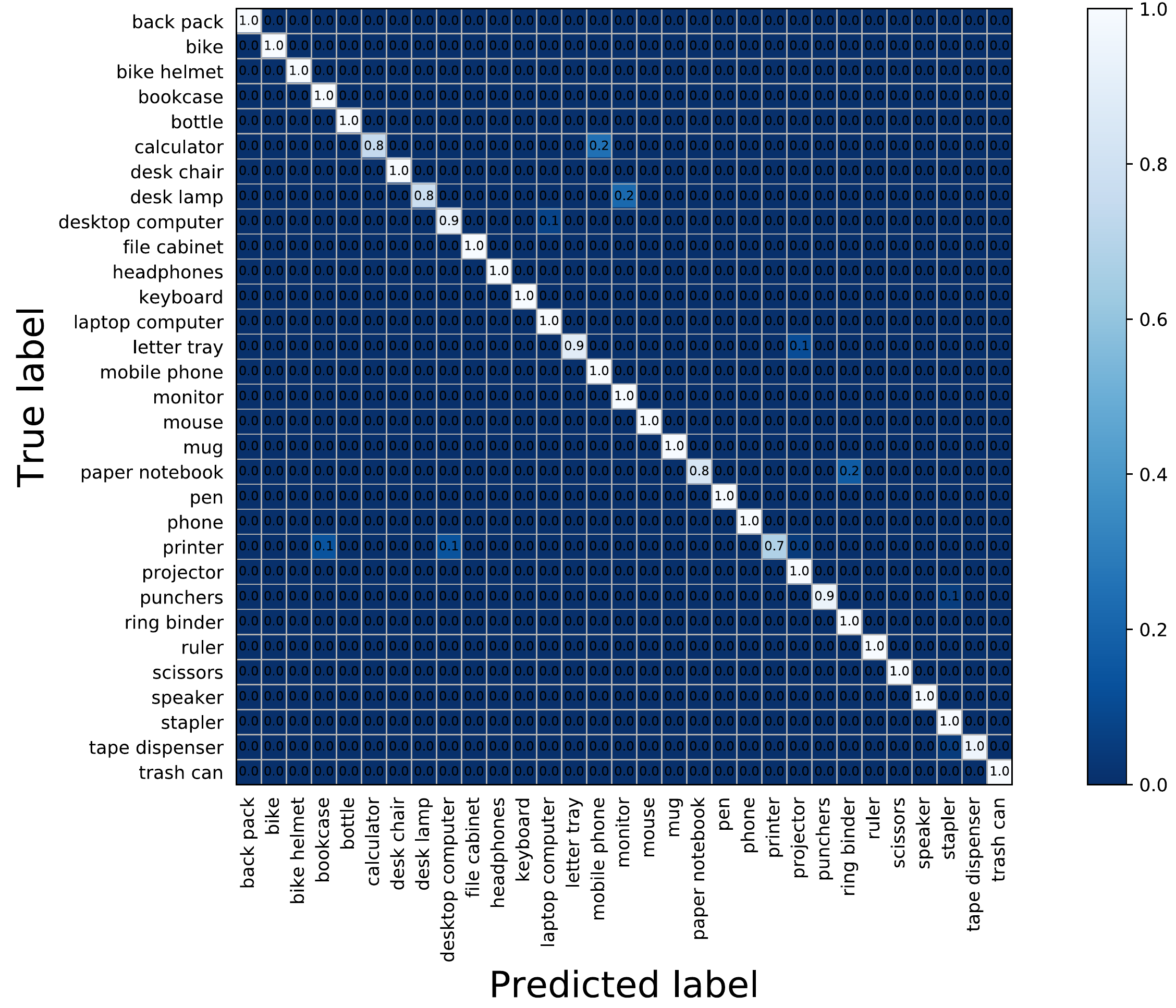}}
  \subfigure[CDAN]{\includegraphics[width=0.235\textwidth]{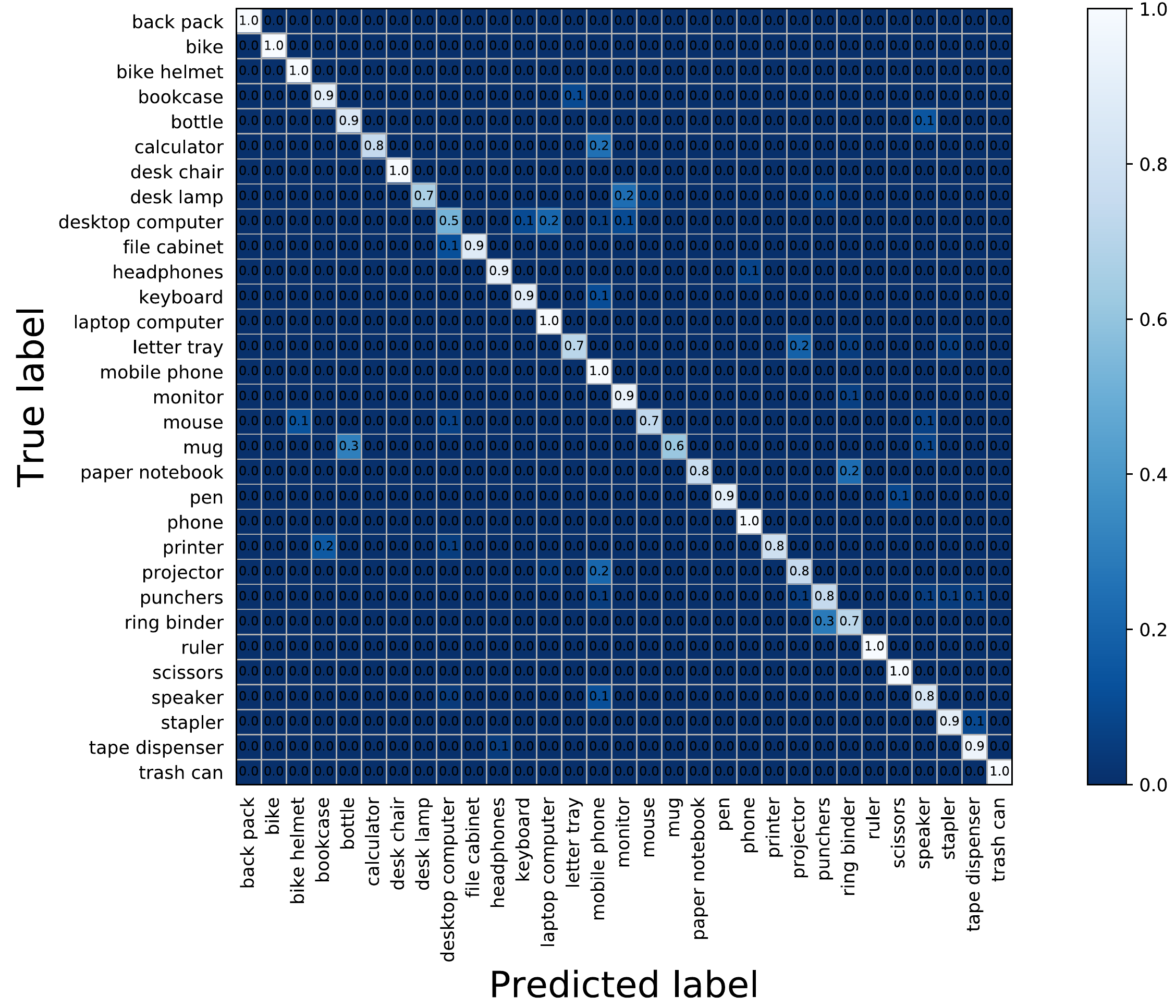}}
  \subfigure[CDAN+SCDA]{\includegraphics[width=0.235\textwidth]{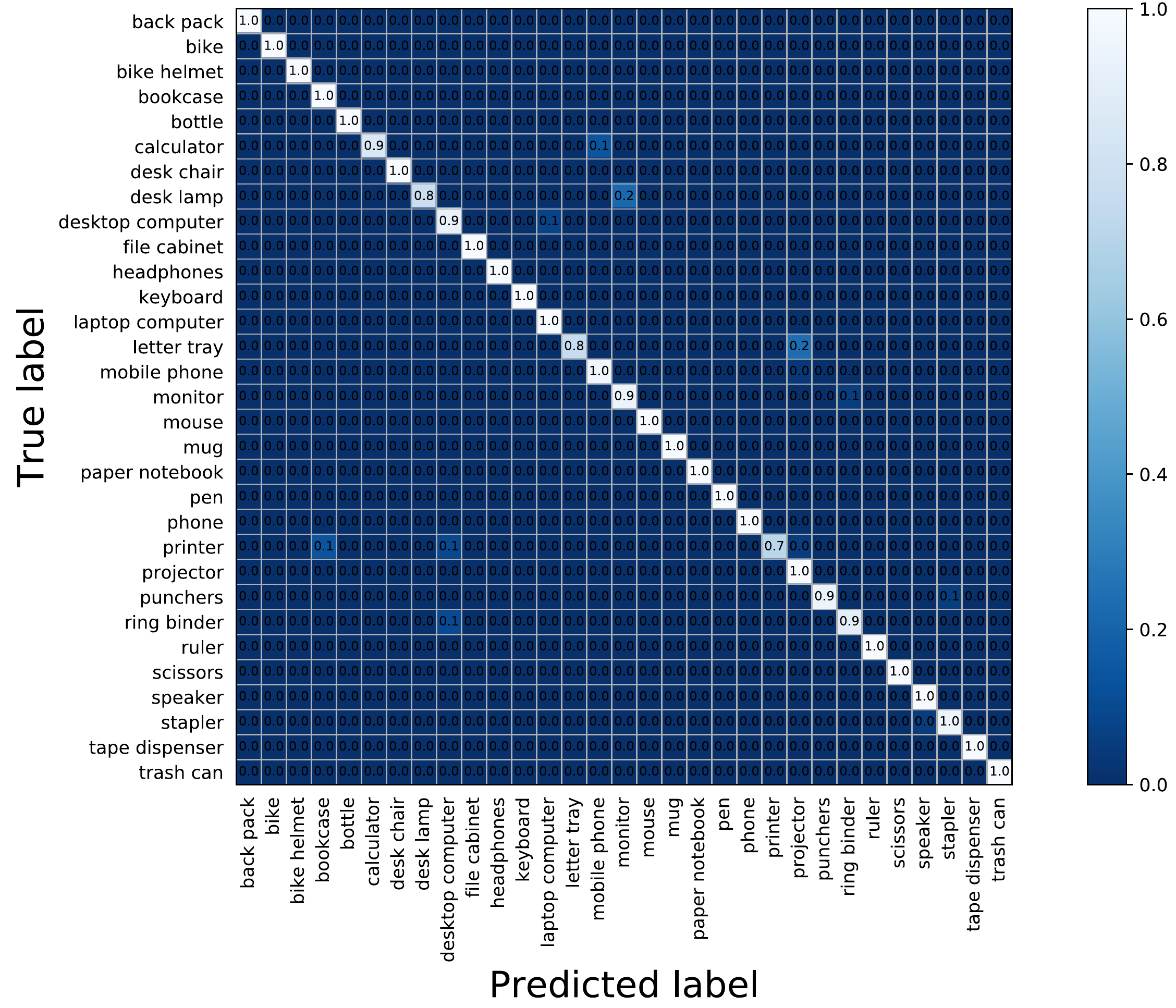}}
  \caption{The confusion matrices on target domain of different methods on the task A $\rightarrow$ D of Office-31. (Zoom in for clear visualization.)}
  \vspace{-4mm}
  \label{Fig_ConfusionMatrix}
\end{figure*}

\textbf{Visual Explanations for Semantic Concentration.} In this section, we utilize the visualization technique in \cite{Grad-CAM} to visualize which region SCDA has concentrated on in the adversarial process, which is shown in Fig. \ref{Fig_Attention_Experiment}. We can observe that the concentration on irrelevant regions significantly increases after maximizing the prediction distribution discrepancy loss $\mathcal{L}_{PDD}$, and then, the features of irrelevant/principal regions are suppressed/accentuated by minimizing the discrepancy, which verifies the aforementioned micro explanations. Besides, the final results demonstrate that our method indeed achieves the semantic concentration for the critical parts in image classification.

\textbf{Anti-jamming Ability Test.} Since our method aims to suppress the features of irrelevant semantics and accentuate the features of principal parts, we conduct this experiment to test its anti-jamming ability by adding Gaussian noises with zero-mean to a batch of randomly selected input images and then testing the sensitivity of different methods as~\cite{sensitivity}. The results are illustrated in Fig.~\ref{Fig_Sensitivity}. It can be clearly observed that the sensitivities of SCDA, CDAN+SCDA and MDD+SCDA (dashed lines) are much smaller and also grow more slowly compared to the corresponding baseline methods (solid lines). Such phenomenon reveals that SCDA can significantly suppress the features of irrelevant noises, further proving the superiority of SCDA.

\textbf{Confusion Matrix.} The confusion matrices of different methods are given in Fig.~\ref{Fig_ConfusionMatrix}. For ResNet-50 and CDAN, there exist numerous wrong predictions appearing in the off-diagonal, e.g., most samples of ``mug" are misclassified into ``bottle". By contrast, we can clearly see quantitative improvements of SCDA and CDAN+SCDA, the reason of which can be explained as the pair-wise adversarial alignment in each class leads to more compact features and thus reduces the class confusion. The encouraging results further show the advantages of SCDA either as an independent method or as a regularizer integrated into existing methods.

\begin{figure}[htbp]
  \centering
  \subfigure[ResNet-50]{\includegraphics[width=0.136\textwidth]{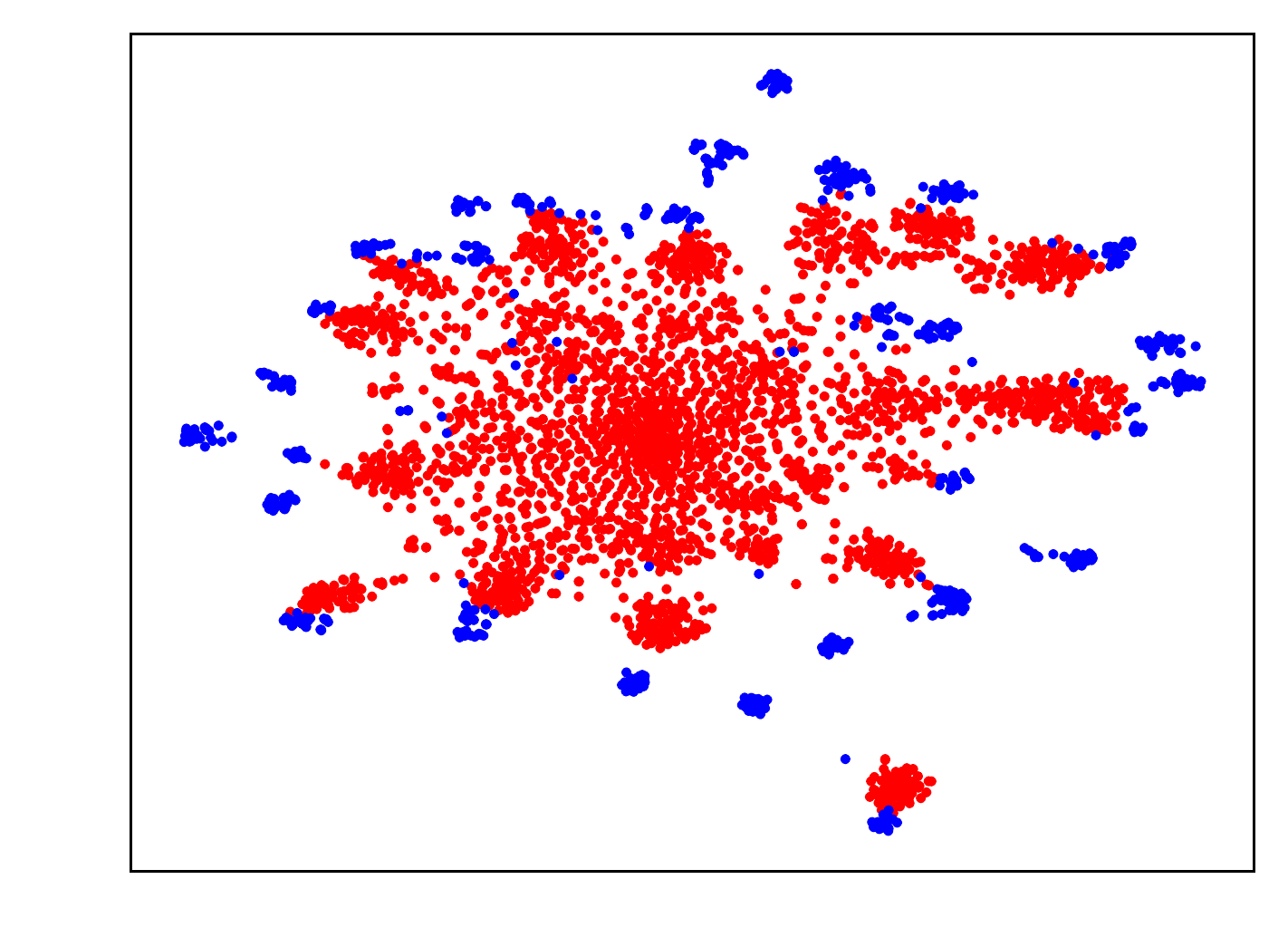}}
  \subfigure[CDAN]{\includegraphics[width=0.136\textwidth]{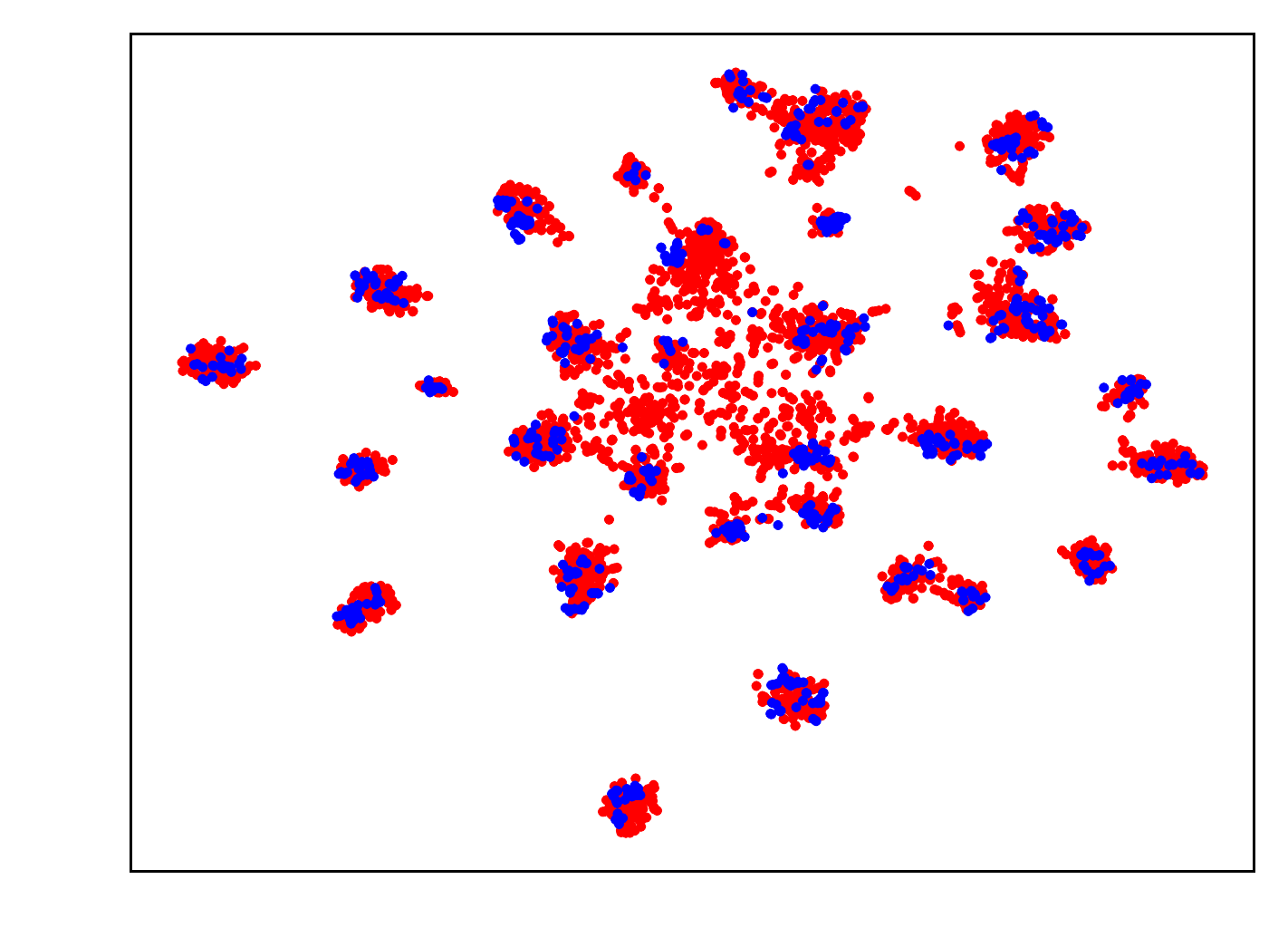}}
  \subfigure[MDD]{\includegraphics[width=0.136\textwidth]{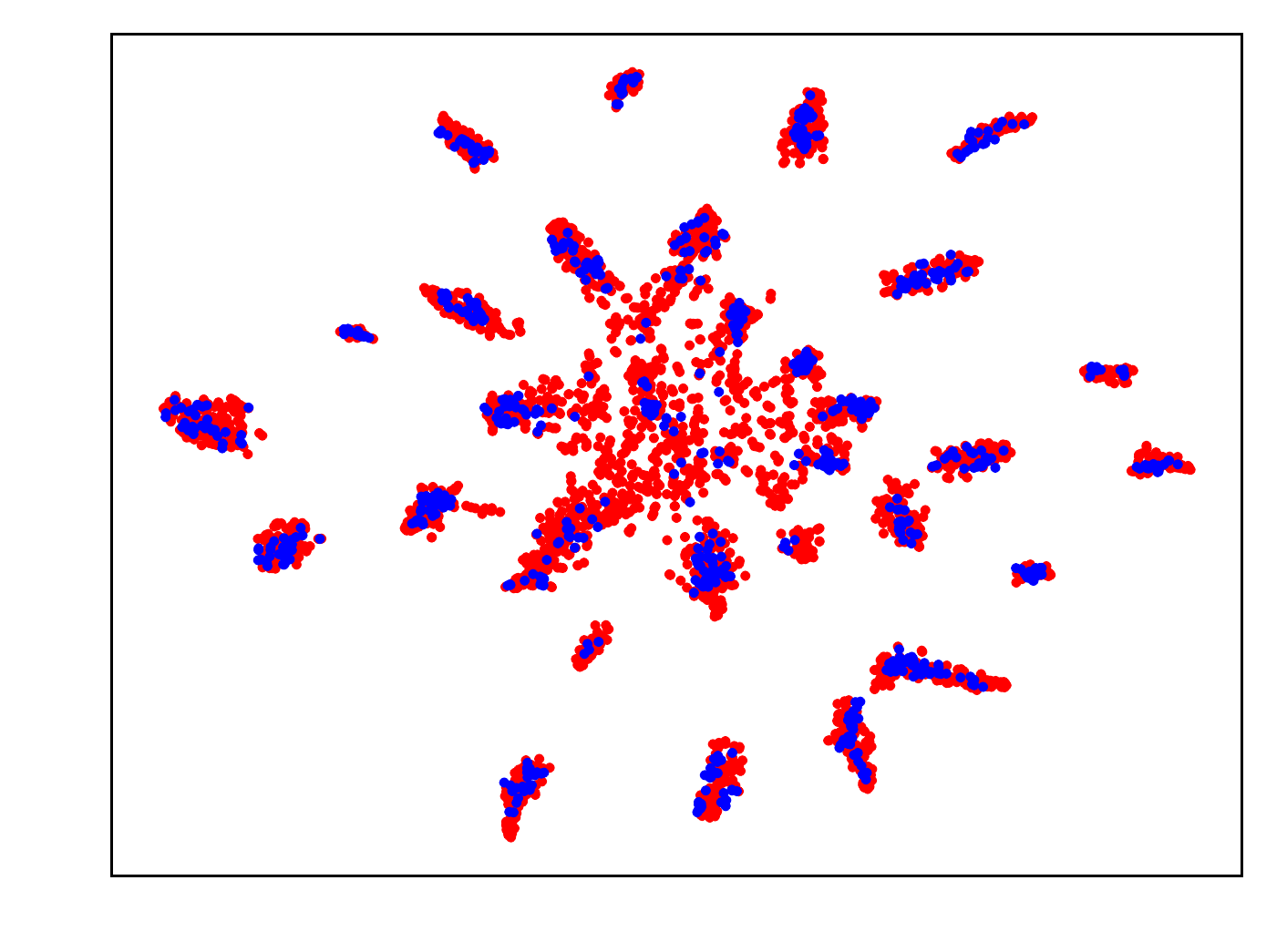}}\vspace{-3mm}
  \subfigure[SCDA]{\includegraphics[width=0.136\textwidth]{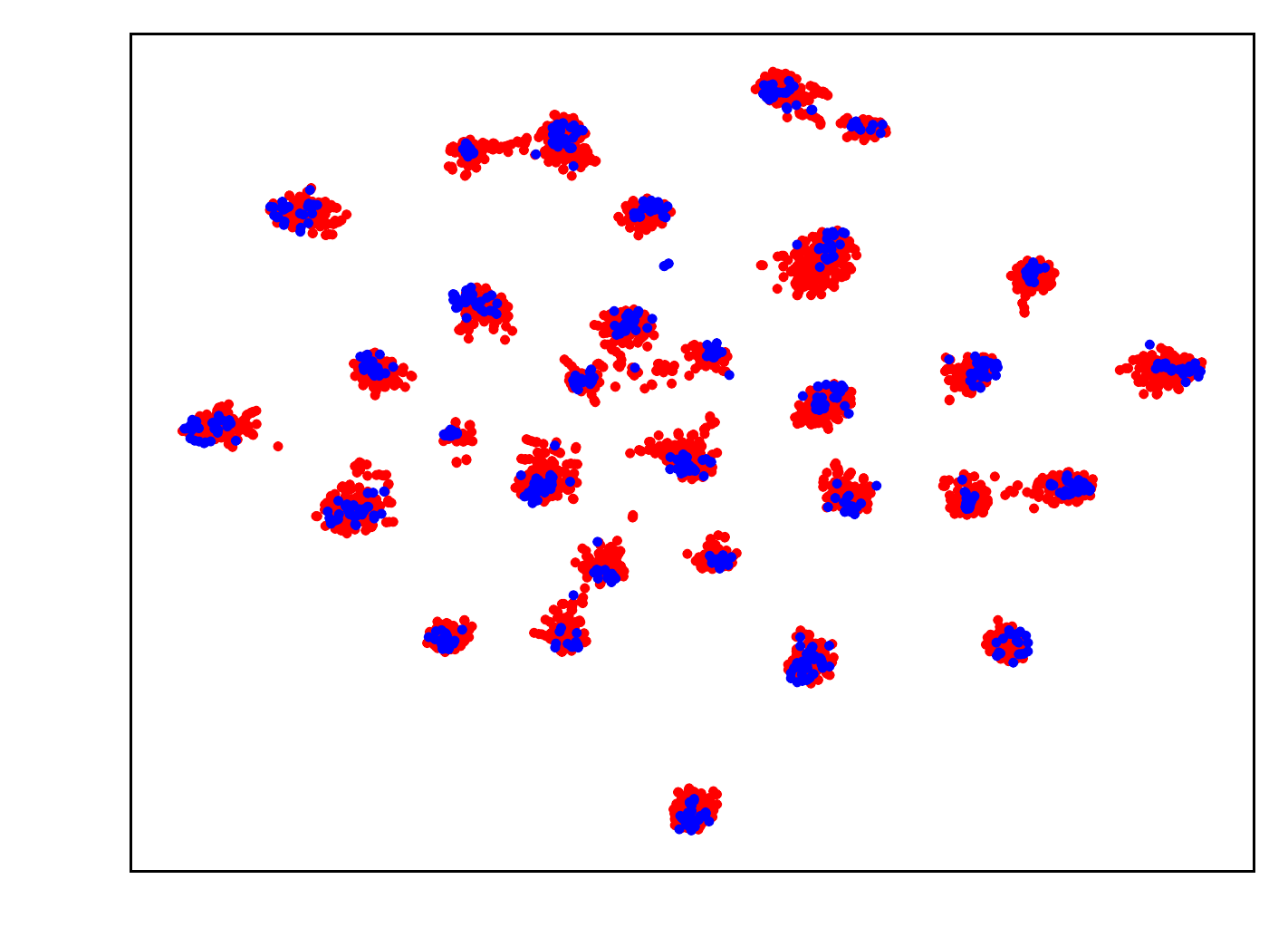}}
  \subfigure[CDAN+SCDA]{\includegraphics[width=0.136\textwidth]{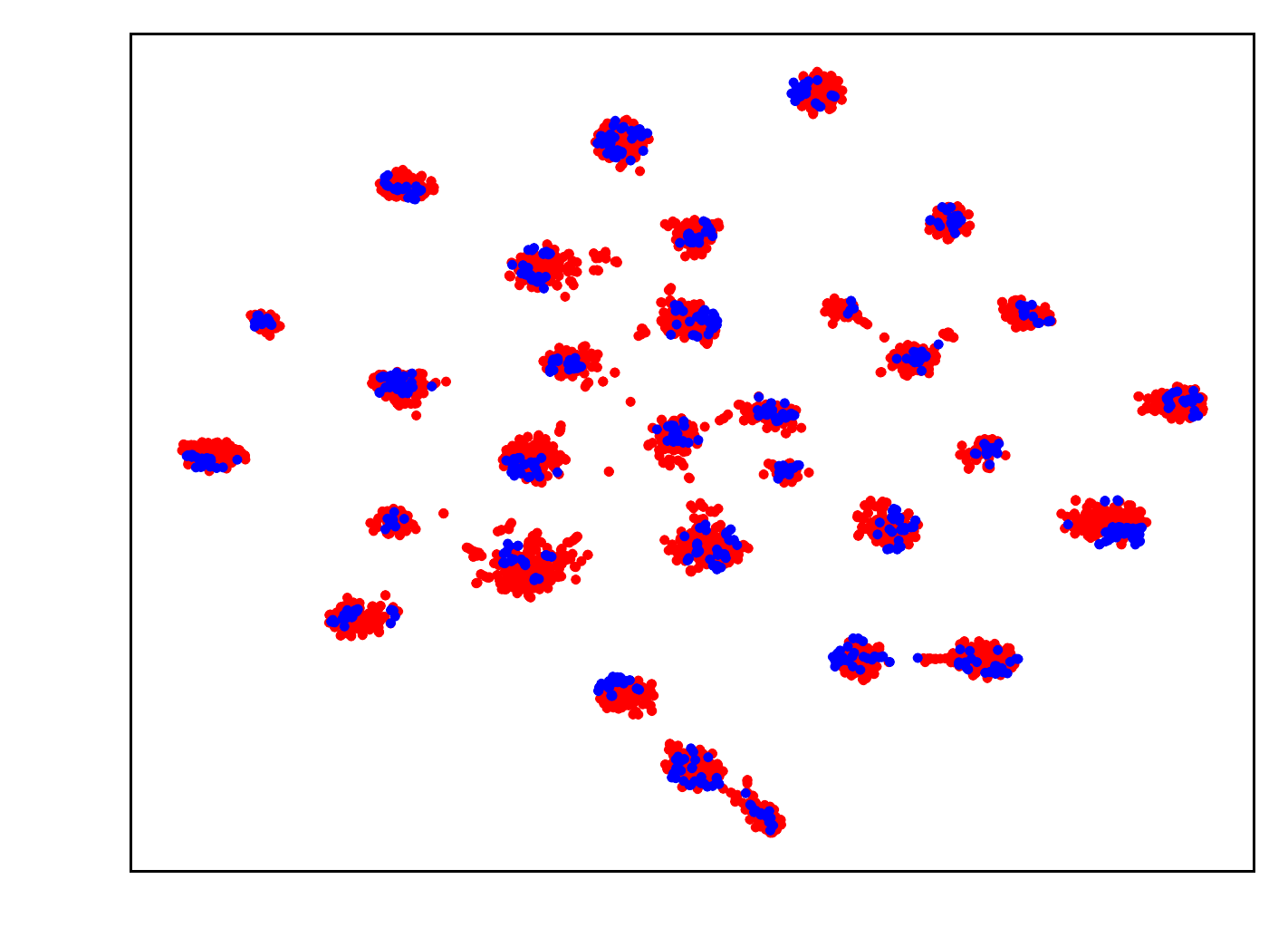}}
  \subfigure[MDD+SCDA]{\includegraphics[width=0.136\textwidth]{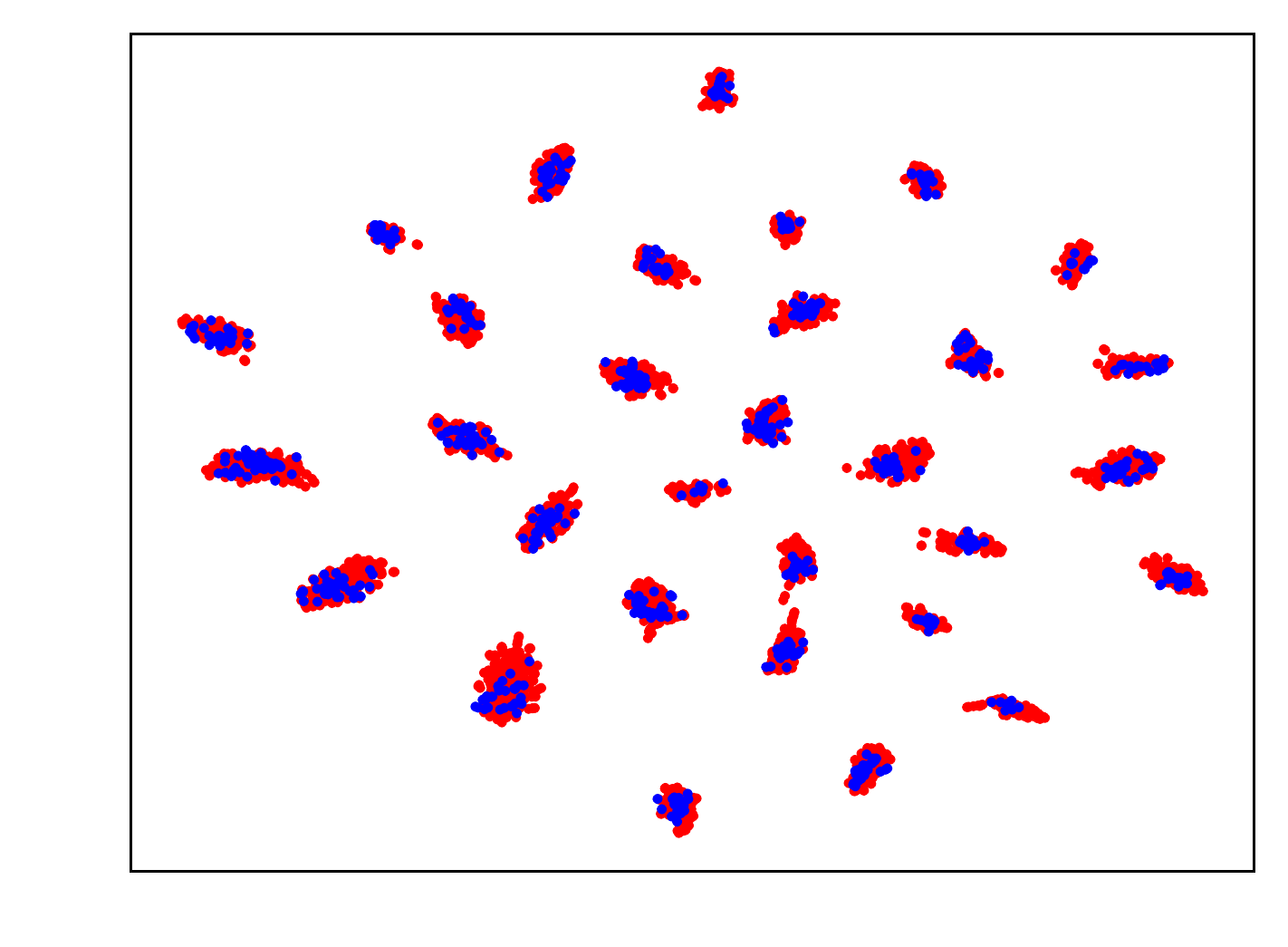}}
  \caption{The visualization of features learned by different methods on the task W $\rightarrow$ A of Office-31. Blue and red dots represent source and target features, respectively.}
  \vspace{-4mm}
  \label{Fig_tSNE_Experiment}
\end{figure}

\textbf{t-SNE Visualization.} Fig. \ref{Fig_tSNE_Experiment} visualizes the feature representations learned by  ResNet-50, CDAN, MDD, SCDA, CDAN+SCDA and MDD+SCDA with t-SNE~\cite{tsne}. We can clearly see that target data are not aligned well with source data using original methods, while SCDA can learn highly discriminative features and keep clear boundaries.

\textbf{Parameter Sensitivity.} Fig. \ref{Fig_ParameterSensitivity_curve}, \ref{Fig_ParameterSensitivity_A2D} and \ref{Fig_ParameterSensitivity_A2W} show the sensitivity of SCDA to temperature $T$, threshold $\epsilon$ and two loss trade-offs $\alpha_0$ and $\beta$ on tasks A $\rightarrow$ D and A $\rightarrow$ W. The results in Fig. \ref{Fig_ParameterSensitivity_A2D} and \ref{Fig_ParameterSensitivity_A2W} show that SCDA is not that sensitive when $\alpha_0 \in \{0.5, 0.75, 1.0\}$ and $\beta \in \{0.1, 0.15\}$. In Fig. \ref{Fig_ParameterSensitivity_curve}, SCDA is not sensitive to $T$, but sensitive to $\epsilon$ (with $\epsilon = 0.8$ working best). Because unreliable pseudo labels will confuse the pairing if $\epsilon$ too small, and too large $\epsilon$ will lead to the insufficient knowledge transfer.
\section{Conclusion}

In this paper, we propose Semantic Concentration for Domain Adaptation (SCDA) to accentuate the features of principal parts and suppress the features of irrelevant semantics via the pair-wise adversarial alignment on the prediction space within source domain and across domains. Orthogonal to most DA methods, SCDA can be easily integrated as a regularizer to bring further improvements. Extensive experimental results verify the efficacy of SCDA.

{\small
\bibliographystyle{ieee_fullname}
\bibliography{Reference_ICCV2021}
}

\end{document}